\documentclass[]{fairmeta}

\DeclareTextFontCommand{\textbf}{\bfseries}

\newcommand{\datasetname}{\textsc{OmnilingualGAIA2}\xspace}
\newcommand{\gaiatwo}{\textsc{GAIA2}\xspace}
\newcommand{\gaia}{\textsc{GAIA}\xspace}

\newcommand{\passk}[1]{$\text{pass}@#1$}
\newcommand{\passallk}[1]{$\text{pass}^{\text{all}}_{#1}$}

\newcommand{\model}[1]{\textsc{#1}\xspace}
\newcommand{\lang}[1]{\texttt{#1}}

\newcommand{\claude}{\model{Claude-4.7-Opus}}
\newcommand{\gptfive}{\model{GPT-5.4}}
\newcommand{\gemini}{\model{Gemini-3.1-Pro}}
\newcommand{\gemma}{\model{Gemma-4-31B-Instruct}}
\newcommand{\qwenthreesixathreeb}{\model{Qwen-3.6-35B-A3B}}
\newcommand{\qwenthreesixdense}{\model{Qwen-3.6-27B}}
\newcommand{\gptoss}{\model{GPT-OSS-120B}}
\newcommand{\kimi}{\model{Kimi-2.6}}

\newcommand{\qwenthreefive}{\model{Qwen-3.5}}

\newcommand{\llamareference}{\model{Llama-3.3-70B-Instruct}}

\ifdefined
\fi

\title{\datasetname: Evaluating the Multilingual Gap in Frontier AI Agents}

\author[]{Andrea Caciolai}
\author[]{Pere-Llu\'is Huguet Cabot}
\author[]{Chierh Cheng}
\author[]{Albert Ventayol-Boada}
\author[]{Gabriel Mejia Gonzalez}
\author[]{Christophe Ropers}
\author[]{Lucas Bandarkar}
\author[]{Sebastian Ruder}
\author[]{Darlene Sakakihara}
\author[]{Elliot Yun}
\author[]{Pierre Andrews}
\author[]{Gr\'egoire Mialon}
\author[]{Romain Froger}
\author[]{Marta R. Costa-juss\`a}

\affiliation[]{Meta Superintelligence Labs}

\abstract{Agentic benchmarks aim to measure how well AI agents plan, search, execute, and recover within realistic multi-tool environments, but they are almost exclusively in English. 
As AI agents are globally deployed to a linguistically diverse user base, whether agentic competence measured in English transfers to other languages remains an open question. 
We introduce \datasetname, a machine-translated expansion (with partial human-expert validation) of the \gaiatwo agentic benchmark, covering ten target languages spanning five writing systems, paired with a localised and human-calibrated multilingual verifier. 
Evaluating seven frontier and open-weight agents, we find a universal cross-lingual gap of 8.8–18.4 \passk{3} points that is agent-asymmetric in magnitude, concentrates on tool-orchestration rather than quantitative reasoning, and does not close with model scale. 
A stratified error attribution decomposes the gap as predominantly model-driven (55\%), with a bounded translation-contamination floor of only 6.4\% of scenario–language pairs. 
Human-expert linguistic analysis further identifies morphological cue loss and amplified ambiguity as the primary failure mechanisms in non-Latin-script languages. 
Our results argue that multilingual agentic evaluation must become a standard part of the reporting protocol for globally deployed agents.}

\date{\today}
\correspondence{\email{costajussa}@meta.com}

\metadata[Dataset]{\url{https://huggingface.co/datasets/facebook/omnilingual-gaia2}}
\metadata[Code]{\url{https://github.com/facebookresearch/meta-agents-research-environments}}
\usepackage{times}
\usepackage{latexsym}
\usepackage[T1]{fontenc}
\usepackage[utf8]{inputenc}
\usepackage{microtype}
\usepackage{inconsolata}
\usepackage{CJKutf8}
\usepackage{graphicx}
\usepackage{booktabs}
\usepackage{longtable}
\usepackage{multirow}
\usepackage{amsmath}
\usepackage{amssymb}
\usepackage{xcolor}
\usepackage{subcaption}
\usepackage{xspace}
\usepackage{pdflscape}
\usepackage[colorinlistoftodos, textsize=small]{todonotes}

\begin{document}

\maketitle

\section{Introduction}
\label{sec:intro}

AI agents---systems that plan, invoke tools, and act autonomously within an environment to pursue a user's goals \citep{wooldridge1995intelligent, franklin1996agent, russell2010artificial}---have moved from objects of research curiosity to real products deployed and used by hundreds of millions of people worldwide, thanks to the rapid advancements of Large Language Model (LLM) capabilities~\citep{wang2024survey, xi2025rise, yang2025adoption, staufer2026index}. 
Those users do not all speak English, yet our ability to measure whether these agents actually work has barely kept pace with their deployment, and least of all beyond English. 
Even in English the ground is unsteady: developers who anticipated a $24\%$ speedup from AI tools were instead slowed by $19\%$~\citep{lobentanzer2026}, and observed agent task coverage remains a fraction of what has been hypothesised~\citep{massenkoffmccrory2026labor,johnstonholtz2026shift}. 
Whether the competence we certify in English survives once an agent must plan, search, and act in another language is, at the time of writing, largely unmeasured.

In English, this gap is being rapidly closed, with considerable work invested in agentic benchmarks that probe how well these systems plan, search, and use tools~\citep{mialon2023gaia,qin2024toolllm,xie2024osworld,patil2025berkeley,deng2025swe,barres2025tau,merrill2026terminal}. 
Beyond English, evaluation has barely followed: for frontier agentic systems\footnote{See for instance \href{https://www.anthropic.com/claude-opus-4-7-system-card}{System Card: Claude Opus 4.7}}, multilingual assessment remains largely confined to single-turn, knowledge-based question answering datasets such as \textsc{Global-}MMLU~\citep{singh2025global}, when it is performed at all\footnote{No multilingual performance reported in \href{https://openai.com/index/gpt-5-4-thinking-system-card/}{GPT-5.4 Thinking System Card}}.

To help close this gap, we present \textbf{\datasetname}, a high-quality machine-translated expansion of \gaiatwo~\citep{froger2026gaia2}, paired with a multilingual verifier and a frontier multilingual agentic leaderboard. Our contributions are:
\begin{enumerate}
  \item \textbf{A multilingual agentic benchmark.} We release \datasetname, a
  machine-translated (MT) expansion of \gaiatwo that covers four capabilities, namely adaptability, ambiguity, execution, and search, across ten target languages spanning five writing systems while preserving the executable structure required by the verifier
  (\S\ref{sec:omnigaia}). We select the translation system through a per-language quality evaluation and design a MT pipeline that enforces cross-surface terminology consistency across scenarios (\S\ref{sec:pipeline}). See our language-coverage contribution relative to existing datasets in Appendix~\ref{app:languages}.
  
  \item \textbf{A human-calibrated multilingual verifier.} We localise the
  LLM-as-a-Judge (LLMaaJ) component of the \gaiatwo verifier, and ensure its multilingual quality via calibration on a golden set of human-annotated and MT agent traces (\S\ref{sec:verifier}).
  \item \textbf{A cross-lingual multi-agent leaderboard and empirical characterisation of the gap}. We evaluate a heterogeneous cohort of seven AI agents spanning frontier closed-source systems and open-weights models of various sizes and architectures, and find that: (i) the cross-lingual gap is universal in direction but agent-asymmetric in magnitude (8.8–18.4 percentage points, pp); (ii) the capability ordering (search > execution > adaptability > ambiguity) is stable across languages - translation reshapes the level without reordering difficulty; (iii) agents shift their behavioural strategy off English, increasing exploratory reads while reducing state-changing writes; (iv) the cross-lingual gap is concentrated in tool-orchestration and final-response quality, not in quantitative or categorical reasoning; and (v) the gap does not close with model scale within a single family (\S\ref{sec:setup}, \S\ref{sec:results}).
\item \textbf{An error taxonomy and gap attribution showing the gap is predominantly model-driven}. We define a five-category verdict taxonomy and apply a stratified automatic triage protocol that corrects for the selection bias of deterministic-only analysis. The resulting decomposition attributes 55\% of the cross-lingual gap to genuine agent failures, 35\% to translation defects, and 10\% to verifier artefacts (\S\ref{sec:taxonomy},\S\ref{sec:automaticgap}). Extending the analysis to the whole benchmark, we bound MT-induced contamination at only 6.4\% of all scenario–language pairs, establishing that the benchmark predominantly measures model capability rather than translation noise.
\item \textbf{A human-expert linguistic analysis identifying failure mechanisms per error category}. Three native-speaker linguists independently inspect agent traces across the four capabilities and surface qualitative patterns: (i) translation defects — MT amplifies latent English ambiguities and silently drops directional/morphological constraints that disambiguate the correct answer; (ii) verifier artefacts — verifier strictness is language-dependent, stochastic, and can hallucinate rejections of correct content; (iii) agent failures — morphological cue loss (articles, plural marking) in cmn/jpn/ind drives the most consequential over-actions; and (iv) measurement artefacts — action-count grading penalises harmless self-corrections that reach the correct end state, inflating the apparent failure rate (\S\ref{sec:linguisticanalysis}).
  
\end{enumerate}

OmniGAIA2 is part of Meta's broader Omnilingual effort to extend AI capability beyond the handful of high-resource languages that dominate current systems, alongside Omnilingual ASR \citep{omnilingualasrteam2025omnilingualasropensourcemultilingual} for speech recognition across 1,600+ languages and Omnilingual MT \citep{omnilingualmtteam2026omnilingualmtmachinetranslation} for translation at comparable scale. Where those efforts expand the capability frontier, and BOUQuET \citep{andrews-etal-2025-bouquet} the translation-evaluation frontier, OmniGAIA2 extends evaluation to interactive, tool-using agentic tasks beyond English.
\section{Related Work} \label{sec:related}

\paragraph{Multilingual agentic evaluation.} Efforts to move agentic evaluation beyond English have so far been narrow along at least one axis: the set of languages, the class of agents evaluated, or the fidelity of the verifier under translation. In the web-agent setting, X-WebAgentBench~\citep{wang-etal-2025-xwebagentbench} translates a subset of navigation tasks into several languages, while Ticket-Bench~\citep{ticketbench2025} regionalises a customer-service function-calling benchmark across six European locales. On the tool-calling sub-capability alone, MLCL~\citep{luo2026lostexecutionmultilingualrobustness} isolates cross-lingual robustness of API selection and argument copying, decomposing the multilingual gap into query-comprehension and parameter-value-mismatch components. MASSIVE-Agents~\citep{kulkarni:2025}, which covers 52 languages, was created by cleaning the original MASSIVE dataset and then reformatting it for evaluation within the Berkeley Function-Calling Leaderboard (BFCL) framework. MAPS~\citep{hofman-etal-2026-maps} pairs a ten-language agent suite with a security-focused analysis, and TelcoAgent-Bench~\citep{bariah2026telcoagentbenchmultilingualbenchmarktelecom} evaluates multilingual telecom agents in English and Arabic, finding that performance gaps widen in unconstrained bilingual settings---a pattern consistent with our own findings across ten languages. Closest to our setting, three recent efforts translate structurally identical benchmarks across languages while preserving their executable verifiers: SEATauBench~\citep{nguyen2026seataubench} adapts $\tau^{2}$-Bench into five Southeast Asian languages, PolyWorkBench~\citep{li2026polyworkbench} constructs a native multilingual long-horizon agent benchmark from scratch, and our most direct sibling, GAIA-v2-LILT~\citep{kim2026gaiav2liltmultilingualadaptationagent}, localises \gaia{} across five target languages through a translate-then-adapt pipeline that modifies task semantics to fit the locale. \datasetname{} instead expands by machine translation: unlike synthetic-generation approaches (IntellAgent~\citep{levi2025intellagentmultiagentframeworkevaluating}, TaskCraft~\citep{shi2025taskcraftautomatedgenerationagentic}, AgentSynth~\citep{xie2026agentsynthscalabletaskgeneration}), MT-based expansion preserves an existing verifier's executable structure while scaling linguistically, following the hybrid MT-then-expert-review pipeline of MMLU-ProX~\citep{xuan-etal-2025-mmlu} at 29 languages. Benchmark validity grounds our analysis: the Agentic Benchmark Checklist (ABC)~\citep{zhu2025establishingbestpracticesbuilding} documents systematic outcome- and task-validity failures in widely used agentic benchmarks; we address these by calibrating the verifier against human labels (\S\ref{sec:verifier}) and by decomposing the observed gap into translation-, model-, and verifier-side factors with a bounded contamination floor (\S\ref{sec:linguisticanalysis}). Against this landscape, \datasetname{} is, to our knowledge, the first extension of \gaiatwo{} that \textbf{(i)} covers ten target languages spanning five writing systems (Appendix~\ref{app:languages}) while preserving the executable verifier end-to-end, \textbf{(ii)} reports a paired comparison across a heterogeneous cohort of frontier and open-weights agents under a shared, calibrated cross-lingual judge, and \textbf{(iii)} decomposes the observed gap into translation-side and model-side factors, extending the analytic programme of MLCL from isolated tool calls to full agent trajectories---complemented by a comprehensive human-expert linguistic validation.

\paragraph{Multilingual LLM-as-a-Judge.} Because \gaiatwo{}, and by inheritance \datasetname{}, relies on a model-based verifier to grade the natural-language write actions of agents, the cross-lingual reliability of the judge itself is a first-class concern. LLM judging was popularised by \citet{zheng2023judging} and inherits from prompt-based translation-quality scoring in the MT community~\citep{kocmi2023gemba,cometkiwi2022,metricx2025}; subsequent audits catalogue position, verbosity, and self-preference biases~\citep{panickssery2024selfrecognition,koo2024benchmarking} that are already problematic in the monolingual setting. In the multilingual setting these effects compound: METAL~\citep{hada-etal-2024-metal} and MM-Eval~\citep{son2024mmeval} document sharp drops in judge-human agreement on lower-resource languages, M-RewardBench~\citep{gureja-etal-2025-rewardbench} reports the analogous collapse for multilingual reward models, BabelJudge~\citep{babeljudge2026} extends the analysis to full agent trajectories and finds order-consistency near chance on some languages, the Coin-Flip-Judge study~\citep{coinflipjudge2026} quantifies reliability floors under adversarial pairings, and \citet{zhang2026dibjudge} isolate a distinct \emph{translationese} bias whereby judges systematically prefer back-translated over human-authored responses. \citet{dogruoz2026challenges} synthesise these findings into concrete recommendations for multilingual judging practice. Our verifier design (\S\ref{sec:verifier}) inherits the \gaiatwo{} LLMaaJ scaffold but explicitly recalibrates it against human-annotated multilingual agent traces, quantifying inter-rater agreement with standard nominal-scale statistics~\citep{cohen1960kappa,artstein-poesio-2008-inter,landis1977measurement,krippendorff2019alpha} and reporting language-conditional confidence intervals on the resulting leaderboard.

\section{\datasetname}
\label{sec:omnigaia}

\subsection{Source benchmark}
We build on \gaiatwo~\citep{froger2026gaia2}, an agentic benchmark in which an
agent operates within a simulated environment equipped with tools. In particular, we build on the native \emph{Mobile} environment, comprised of \emph{apps} such as \emph{Email} and \emph{Calendar}, as well as of a set of corresponding collections of \emph{app states}, aka \emph{universe}. Each universe identifies a synthetic \emph{persona} and a snapshot of their digital life, with the app states containing synthetic user data such as emails or calendar events. Each scenario then specifies a user task in this environment and a minimal (logically and/or temporally ordered) set of actions that the agent must have undertaken, including the final answer. The actual actions (\emph{events}) undertaken by the agent are then compared to these \emph{oracle events}, and in particular the ones containing free-form text (e.g. the final answer) are subject to LLMaaJ verification, to declare the scenario a success or a failure.

The full \gaiatwo capability taxonomy comprises seven splits: Execution, Search, Ambiguity, Adaptability, Time, Noise and Agent2Agent~\citep{froger2026gaia2}. We initially design \datasetname to cover four capabilities, namely \textbf{Adaptability}, \textbf{Ambiguity}, \textbf{Execution} and \textbf{Search}, while reserving the extension to the other capabilities for future work, as those are the ones, at the time of writing, farthest away from saturation by frontier agentic systems even in English, and we posit multilingual performance to lag behind even more.

\subsection{Translating the benchmark}
\label{sec:pipeline}

\paragraph{Translation pipeline}

Translating a \emph{verifiable} agentic task is materially harder than translating a static QA pair: the translated environment must remain internally consistent so that the verifier still admits exactly the intended solution. To construct \datasetname we translate into each target language the language-dependent surface of every scenario: the user task description, the natural-language portion of the universes, as well as of the oracle events (\emph{text spans}). 
The executable structure++ of the scenario, such as identifiers and timestamps, is left untouched (\emph{frozen spans}).

\begin{figure*}[htb]
  \centering
   \includegraphics[width=0.95\linewidth]{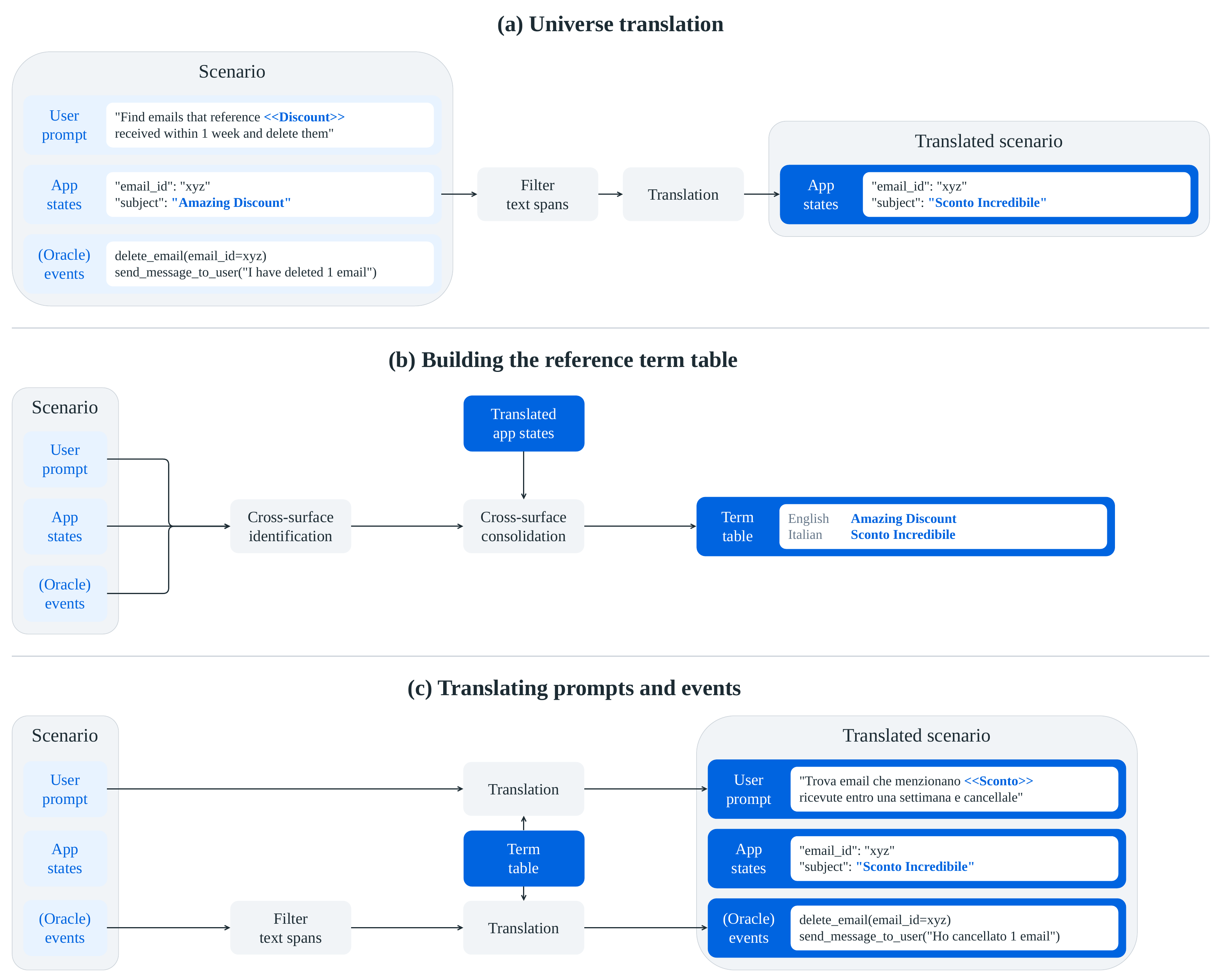}
  \caption{\textbf{The \datasetname{} translation pipeline.} 
  Each \gaiatwo{} scenario is translated in three stages.
  \textbf{(a) Universe Translation} The language-dependent \emph{text spans} of every surface---the user prompt, the text span of the app-states of the universes, and the (oracle) events---are rendered into the target language by the LLM translator, in calls batched per app.
  \textbf{(b) Building the reference term table} A second LLM pass identifies the entities that recur across surfaces and consolidates them into a reference \emph{term table}: a reference rendering of each shared entity.
  \textbf{(c) Translating prompts and events} The term table is applied back across the assembled scenario, substituting any residual or divergent source-language span so that every surface refers to each entity identically---the cross-surface consistency the verifier requires to admit exactly the intended solution.}
  \label{fig:pipeline}
\end{figure*}

The pipeline proceeds in three stages, illustrated in Figure~\ref{fig:pipeline}. Since \gaiatwo defines multiple scenarios over the same universe, the pipeline first aggregates the universes and translates their text spans in calls batched per app. From these translated surfaces a reference \emph{term table} is built---primarily derived from the already-translated universes and augmented by a per-scenario extraction pass---by identifying and consolidating the entities that recur across surfaces; this table enforces cross-surface translation consistency, which is crucial to avoid constructing unsolvable tasks that mention non-existing entities. Task descriptions and oracle-event text spans are then translated into the target language with the term table constraining their rendering, and a final validation pass sweeps the assembled scenario and substitutes any residual source-language span against the table, so that the translated environment stays consistent with what the verifier expects.

\paragraph{Translation system.} We rely on BOUQuET\footnote{\href{https://huggingface.co/spaces/facebook/bouquet}{facebook/bouquet}}~\citep{andrews-etal-2025-bouquet} to select \gemma~\citep{gemmateam2025gemma3technicalreport} as our translator system, as the top-ranking open-license translator system across all the languages we target. We measure translation quality by pairwise LLM-as-judge comparison of each candidate rendering against a reference translation, adjudicated by two independent judges under position randomisation and majority voting; a per-language head-to-head against the other shortlisted systems on this measure confirms this choice.  
Furthermore, we investigate the effectiveness of a review and post-editing pass with a different model, which proves a near-no-op: it rewrites fewer than $9\%$ of fields, with edits that are almost entirely stylistic, with no measurable impact on translation quality while roughly doubling per-scenario latency. See  Appendix~\ref{app:translator} for additional details.

\subsection{Multilingual Verifier}
\label{sec:verifier}

\gaiatwo evaluates every state-changing write action that the agent performs in the environment against oracle annotations via a multi-step \emph{verifier} employing an LLMaaJ component for text spans.
Therefore, to properly extend the benchmark into a multilingual setting, besides the user tasks, universe app states, and oracle events, the LLMaaJ component too needs to be localised, to ensure a fair judgement of non-English scenarios. Here, we use the term \emph{judge} to refer to the LLMaaJ component of the verifier, and describe the work done to localise the judge, and run calibrations to ensure parity to English and correctness on target languages. We summarize the protocol and findings below, and refer the reader to Appendix~\ref{app:judge_calibration} for additional details.

\paragraph{Localising the judge.} The (multilingual) judge quality can be improved along two axes: the \textbf{judge prompt} and the \textbf{judge model}. To evaluate both independently of agent performance, we use a set of human-annotated scenario traces from \gaiatwo, each carrying a ground-truth pass/fail label. The same set is translated with our pipeline (\S\ref{sec:pipeline}) to produce a cross-lingual counterpart, enabling evaluation under both source and target languages on identical trace scaffolding.

Starting the investigation with the judge prompt, we find English-specific assumptions that make these prompts unsound on translated content; e.g. the content checker is primed exclusively with English few-shot examples, so a faithful non-English write action can be judged non-compliant for reasons unrelated to its correctness. \datasetname therefore ships a set of localised judge prompts that remove this priming and generalise the natural-language sub-checks across target languages; the calibration below confirms that this localisation leaves English agreement intact while carrying over to the target languages we evaluate.
On the model axis, we require a judge that is open-weight and light-weight enough to score the entire multilingual leaderboard. 
\gptoss\footnote{the configuration the current \gaiatwo reference implementation itself adopts, see \href{https://github.com/facebookresearch/meta-agents-research-environments}{facebookresearch/meta-agents-research-environment}}, meets both: as a mixture-of-experts model it activates only $5.1$B parameters per token~\citep{openai2025gptoss}, roughly an order of magnitude fewer than the dense~\llamareference original paper's reference judge~\citep{meta2024llama33}, hence far fewer FLOPs per call. 

\paragraph{Calibrating the judge.} 
\label{sec:verifier-calibration}
We compare, on the same human-annotated traces, the four combinations of the two judge models with the two prompt sets, scoring each against the human-majority label with Cohen's $\kappa$, for chance-corrected agreement.

On English, no configuration reproduces the human verdicts measurably better than another: the four agree with the human labels comparably (full-corpus $\kappa$ within $0.71$--$0.73$, rising to $\kappa \geq 0.93$ on the \emph{LLM-touched} slice the judge actually adjudicates). Crucially, the proposed configuration (\gptoss, localised) agrees with the original paper's reference judge on almost all traces, with $\kappa = 0.981$. 
This agreement is preserved under translation across all ten target languages, on the LLM-touched slice Cohen's $\kappa$ ranges from $0.886$ (\lang{tur}) to $0.971$ (\lang{eng}, \lang{cmn}).

\clearpage
\section{Experimental Setup}
\label{sec:setup}

\paragraph{Models}
We evaluate seven frontier and open-weight models on \datasetname, selected to span both the closed-source vs open-weights divide and, within open-weights, across dense and mixture-of-experts architectures at varied scale. 
Among proprietary systems, we include \claude~\citep{anthropic2025claude4}, \gptfive~\citep{openai2025gpt5}, \gemini~\citep{google2025gemini3}. 
For open-weight models, we evaluate \gemma~\citep{google2025gemma4}, \kimi~\citep{moonshot2025kimi} and the Qwen~3.6 family in both its mixture-of-experts (\qwenthreesixathreeb) and dense (\qwenthreesixdense) configurations~\citep{qwen2025qwen3}. We also conduct a scale ablation on the latter family.

\paragraph{Agent Harness}
All agents are driven by the \textsc{OpenClaw}\footnote{\url{https://github.com/openclaw/openclaw}} harness, with the exception of \kimi, which is driven by \textsc{OpenCode}\footnote{\url{https://opencode.ai}}, as it scores near zero when driven by \textsc{OpenClaw} instead.\footnote{We root-cause this to the harness dropping its native streaming \texttt{reasoning\_content} and \texttt{tool\_calls}.}.

\paragraph{Generation Settings.} We serve the open-weight agents with a default $131$K context window, re-scoring with an augmented context window (using the model's maximum) those rollouts that fail merely because its trajectory was truncated\footnote{Truncation is concentrated in three non-Latin languages, affecting up to $20\%$ of \lang{cmn}, $34\%$ of \lang{jpn} and $44\%$ of \lang{hin} rollouts.}.
For open-weight models, we enable thinking mode and otherwise adopt each model's default generation configuration.
For proprietary models, we use each provider's default reasoning-effort configuration. To verify that this choice does not drive our results, we run a controlled, smaller scale reasoning-effort ladder and find the cross-lingual gap persists, largely undiminished, at the maximum tested effort (Appendix~\ref{app:effort}).

\paragraph{Metrics.} We report \passk{3} as our main metric: a scenario counts as solved if any of three independent runs passes the verifier, for direct comparability with~\gaiatwo~\citep{froger2026gaia2}.  We also monitor \passk{1} and \passallk{3}.

\section{Results}
\label{sec:results}

\subsection{Leaderboard}
\label{sec:results-headline}

Table~\ref{tab:headline} reports pooled \passk{3} numbers over the four capabilities (adaptability, ambiguity, execution, search).

\begin{table*}[t]
  \centering
  \small
  \begin{tabular}{l ccccc}
    \toprule
    Agent & Exec. & Search & Adapt. & Ambig. & Avg. \\
    \midrule
    & \multicolumn{5}{c}{\textbf{English}} \\
    \cmidrule(lr){2-6}
    \claude              & \textbf{88.1} & \textbf{95.6} & \textbf{75.6} & \textbf{71.6} & \textbf{82.7} \\
    \gptfive             & 81.2 & 92.4 & 58.8 & 41.4 & 68.4 \\
    \gemini              & 66.2 & 86.9 & 51.2 & 44.4 & 62.2 \\
    \addlinespace[2pt]
    \kimi                & 73.8 & 94.3 & 55.0 & 42.8 & 66.5 \\
    \gemma               & 67.5 & 85.6 & 63.1 & 36.9 & 63.3 \\
    \qwenthreesixathreeb & 68.8 & 86.2 & 48.1 & 23.1 & 56.6 \\
    \qwenthreesixdense   & 77.5 & 79.4 & 61.2 & 30.6 & 62.2 \\
    \midrule
    & \multicolumn{5}{c}{\textbf{Target-language average}} \\
    \cmidrule(lr){2-6}
    \claude              & \textbf{78.8} & 83.0 & \textbf{73.0} & \textbf{59.7} & \textbf{73.6} \\
    \gptfive             & 67.1 & 83.5 & 51.4 & 31.8 & 58.5 \\
    \gemini              & 56.1 & 78.3 & 48.9 & 30.4 & 53.4 \\
    \addlinespace[2pt]
    \kimi                & 56.5 & \textbf{84.8} & 44.5 & 31.1 & 54.2 \\
    \gemma               & 55.2 & 69.5 & 53.4 & 27.5 & 51.4 \\
    \qwenthreesixathreeb & 41.8 & 65.6 & 31.8 & 13.6 & 38.2 \\
    \qwenthreesixdense   & 55.2 & 69.4 & 51.9 & 24.4 & 50.2 \\
    \bottomrule
  \end{tabular}
  \par\smallskip
\caption{%
Per-capability \passk{3} (\%) across agents; best per column in \textbf{bold}. The target-language average pools the ten non-English languages.%
}
  \label{tab:headline}
\end{table*}

The results demonstrate a clear and evident cross-lingual gap between the English baseline and any target-language across all agents. This gap varies in magnitude across models, with \gemini{} at $8.8$ \passk{3} points and \claude{} at $9.1$. On the other end of the range, \qwenthreesixathreeb{} shows a pooled gap of $18.4$. The ordering amongst the seven agents is consistent with \passk{1}, with \gemini{} still showing the smallest pooled gap at $7.7$. 
However, \gemini{} has the lowest English and multilingual performance of the three frontier agents ($62.2$ \passk{3}). Within the open-weights class \kimi{} outperforms \gemini{} in English but lags behind on target languages.

Looking at the \passk{1}, \passk{3} and \passallk{3} numbers (reported in~Appendix~\ref{app:per_language}) we observe how for most agents extra attempts do not recover the loss and the same target-language scenarios fail every time. 
\claude{} is the exception: its gap narrows from $14.0$ at \passk{1} to $9.1$ at \passk{3} yet is widest at \passallk{3} ($17.5$), so off English it still reaches the solution within three attempts, only far less consistently. 
That target-language \passk{3} stays high (e.g.\ $\sim74\%$ for \claude{}, $58\%$ for \gptfive{}) shows these scenarios remain largely solvable off English rather than the benchmark saturating; how much of the residual difficulty reflects genuine model limitations versus translation artefacts is what we disentangle in the attribution analysis later.

Since \gemma{} serves as both the model in the translation pipeline (\S\ref{sec:pipeline}) and one of the evaluated agents, we verify that this leaderboard is not distorted by a \emph{family-match} advantage, i.e. an agent scoring higher on data translated by its own model family. We run a Gemma-vs-Qwen agent-translator ablation over all ten target languages, and find the agent gap invariant to which family translated the evaluation data, with both agents losing comparably on the weaker Qwen translations; full design and per-language results are in Appendix~\ref{app:family-match}.

\subsection{Per-capability breakdown}
\label{sec:analysis-capability}

\begin{figure*}[t]
\centering
\includegraphics[width=\textwidth]{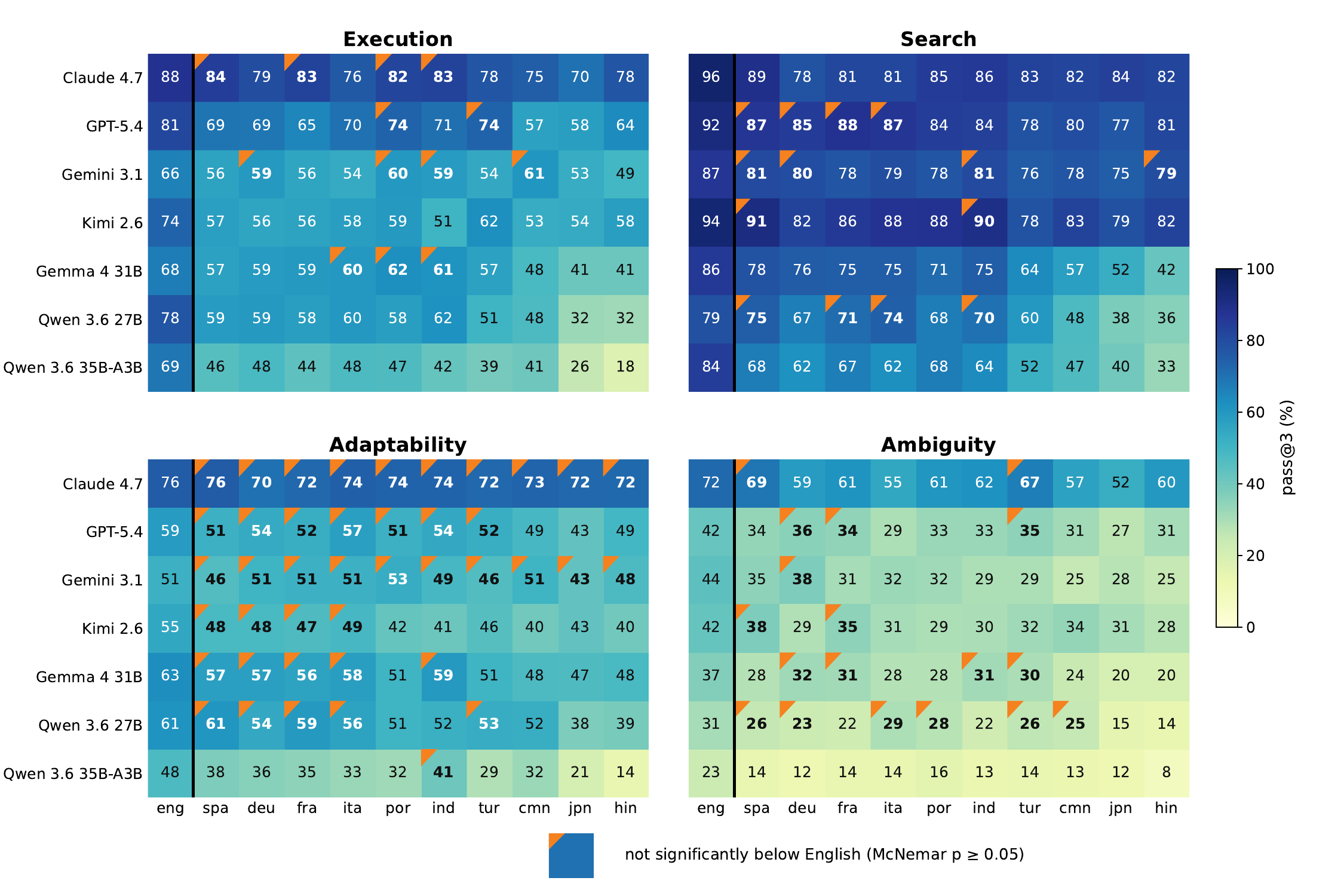}
\caption{Per-capability $\times$ per-language \passk{3} (\%) across agents. 
Cross-lingually robust cells, i.e. target language cells that are \emph{not} significantly different from the corresponding English cell from the same (agent, capability) combination under a paired McNemar test ($p\geq0.05$), are marked with an \textcolor{orange}{orange corner wedge}.}
\label{fig:heatmap_breakdown}
\end{figure*}

We now break down the numbers of Table~\ref{tab:headline} into the four capabilities. 
For each (capability, language, agent) combination we report \passk{3} with $95\%$ CIs and we test the English-vs-target-language difference with a paired McNemar test~\citep{mcnemar1947} on the matched scenario pairs, per capability.
Figure~\ref{fig:heatmap_breakdown} gives the full capability-by-language view at a glance, while the exact numbers, Wilson CIs and per-cell McNemar $p$-values are in Appendix~\ref{app:per_language}.
Two patterns are visible: performance is strongest in English for nearly every (agent, capability) combination, and the drop concentrates on the CJK and Indic languages (\lang{cmn}, \lang{jpn}, \lang{hin}). These three languages notably use non-Latin scripts, and we find that part of the reason is the agents not carefully considering the various forms \textit{entities} may take when calling tools (see Appendix~\ref{script_analysis} for details), mirroring cross-script mistakes in non-agentic tasks \citep{bandarkar2026largereasoningmodelsstruggle}. The frontier agents (\claude, \gptfive) show the most cross-lingually robust behaviour, while the open-weights agents perform significantly below English on almost every target language.

Furthermore, the capability ordering appears stable across languages, with \passk{3} numbers decreasing from search through execution and adaptability to ambiguity, so translation reshapes the \emph{level} of each capability without re-ordering which capabilities are easy.
Search is both the strongest capability and among the more robust, with \kimi{} achieving the highest target-language search of any agent ($84.7$, edging \claude{}'s $83.2$ and \gptfive{}'s $83.1$)
The frontier agents are most cross-lingually robust on \emph{adaptability}: \claude{} and \gemini{} retain it almost intact off English (target-language means of $73.0$ and $48.9$ against English $75.6$ and $51.2$, leading to gaps of $2.6$ and $2.3$ points), so their pooled loss is carried instead by execution and, especially, ambiguity, which is the hardest capability everywhere and the one on which every agent scores lowest off English.

\subsection{Behavioural signatures}
\label{sec:analysis-behaviour}

Comparing the English and target-language distribution of lower-level per-scenario signals, such as agent step count and tool-call composition (which tools are invoked and in what proportion) allows us to zoom in on the agents' behavioural modes that shift under translation: inflated step counts and tool-call retry loops may indicate the agent struggling with a partially-understood instruction, whereas truncated planning or a collapsed tool-call vocabulary
might suggest the agent giving up early rather than exploring. We focus our analysis on two proprietary agents (\claude{}, \gptfive{}) and the strongest open-weights agent (\kimi{}), and find two main behavioural signatures.

\paragraph{Tool-call volume.}
For each scenario we pair its target-language runs with the matching English runs and compare the average number of tool calls. Only \gptfive{} does systematically more work off English: it issues $21.6\%$ more tool calls than in English, an increase present in \emph{every} target language and largest for \lang{hin} ($+41\%$), \lang{spa} ($+37\%$), \lang{tur} ($+23\%$), \lang{deu} ($+22\%$) and \lang{fra} ($+21\%$). \claude{} and \kimi{} do the opposite, issuing slightly \emph{fewer} calls off English on average ($-7.6\%$), with per-language changes mixed in sign (\claude{} from $-23\%$ on \lang{por} to $+7\%$ on \lang{cmn}; \kimi{} from $-17\%$ on \lang{fra}/\lang{ita}/\lang{spa} to $+6\%$ on \lang{hin}).
This extra activity does not, however, explain the regression. For \gptfive{} the scenarios that lose accuracy off English inflate their tool count no more than those that hold it ($+4.8$ vs $+6.5$ calls), so tool volume and success move independently. The one agent whose tool count does track its accuracy is \claude{}, and it runs the \emph{opposite} way: its regressing scenarios issue \emph{fewer} calls, not more ($-11.0$ vs $+3.6$ calls)---a ``give up early'' signature rather than flailing with extra actions.

\paragraph{Tool-strategy composition.}
All three agents also change \emph{which} tools they reach for off English, and this shift is far larger than a cohort-level average suggests. We quantify it with the Jensen--Shannon divergence (JSD) between an agent's English and target-language tool-choice distributions, where higher means a bigger strategy shift. 
Pooled over a whole cohort the shift looks negligible ($0.010$/$0.025$/$0.020$ bits for \claude{}/\gptfive{}/\kimi{}), because opposing shifts in different scenarios cancel out, but measured per matched scenario it is several times larger ($0.087$, $0.109$, $0.116$ bits), and larger still within matched thirds of each trajectory ($0.170$, $0.197$, $0.216$), an approximate per-turn view. 
Two things follow: the pooled figure badly understates how much each agent re-plans its tool use off English; and, measured per scenario, the three agents look much more alike, \gptfive{}'s apparent $2.5\times$ lead over \claude{} shrinks to about $1.3\times$, and \kimi{} re-plans its tool use as much as either frontier agent.
\paragraph{What changes.} The direction of the shift is consistent across all three agents and across languages: off English they spend a larger fraction of their tool budget on \emph{exploratory reads} and a smaller fraction on \emph{state-changing writes}. 
The categories whose share grows most are retrieval and browsing operations: product search and catalogue listing rise by $+3$--$6$\,pp for every agent, with further gains on product-detail lookups and conversation listing. 
The per-tool \emph{falling} side is noisier (dominated by scenario-mix effects), so we summarise the commit side by aggregating over the mutation flag each tool carries: the write-action share of all tool calls falls from $32.1\%$ to $29.9\%$ for \claude{} ($-2.1$\,pp), $30.5\%$ to $28.1\%$ for \gptfive{} ($-2.5$\,pp) and $29.3\%$ to $28.0\%$ for \kimi{} ($-1.3$\,pp). 
In other words, in a non-English scenario all three agents forage longer over the environment and commit later and less decisively -- the behavioural counterpart to the tool-count-mismatch failures of \S\ref{sec:analysis-checker}, where the wrong \emph{number} of side-effectful
calls is the dominant non-English failure signature.

\subsection{Verifier decomposition}
\label{sec:analysis-checker}

To identify \emph{which} verifier checks drive the regression, we label each failed run by the first verifier check it fails: quantitative (e.g. a mismatched count of retrieved emails), categorical (e.g. mismatched attendee-set for a calendar event), textual (e.g. rejected message content from the LLMaaJ), overall tool-count mismatch, or timeout. 
Then, we report each class's \emph{absolute} failure rate, with all runs in the denominator, rather than its share of failures. 
The share-of-failures view is flat across languages and hides the effect; the absolute view localises it. 
The cross-lingual regression is \emph{not} carried by quantitative or categorical reasoning: from English to the non-English average, quantitative-checker failures move by only $+0.04$ percentage points (pp) for \claude{}, $-0.33$\,pp for \gptfive{} (i.e.\ slightly \emph{improving}) and $+0.06$\,pp for \kimi{}, and categorical-checker failures by $+0.08$, $+0.05$ and $-0.12$\,pp.
None of these deltas are significant under a matched-pair McNemar test. 
The regression is instead concentrated in the
tool-count-mismatch checker ($+10.6$\,pp \claude{}, $+5.3$\,pp \gptfive{},
$+5.7$\,pp \kimi{}) and the textual-reply checker ($+6.2$, $+3.4$, $+3.5$\,pp),
which together account for essentially all of the $+17.0$/$+9.3$/$+11.1$\,pp
rise in the overall non-English failure rate. The cross-lingual gap is thus a
tool-orchestration and final-response-quality effect, not a numeric- or
categorical-reasoning effect -- consistent with the ambiguity over-commitment of
\S\ref{sec:analysis-capability} (tool-count) and the reply-quality degradation
discussed in \S\ref{sec:attributionalgap} (textual).

\subsection{Scale ablation: the \qwenthreefive{} size ladder}
\label{sec:analysis-scale}

We vary \emph{scale} within a single family to ask how agentic competence and the cross-lingual gap of \S\ref{sec:analysis-capability} move with model size. 
We evaluate four \qwenthreefive{} sizes: one dense (\model{27B}) and three mixture-of-experts (\model{35B-A3B}, \model{122B-A10B}, \model{397B-A17B}, with $3$/$10$/$17$B active parameters)\footnote{We omit checkpoints smaller than \model{9B}, for which the agent harness does not reliably elicit tool calls, yielding near-zero valid trajectories.}. 
We run the full benchmark suite of $11$-language $\times$ $4$-capabilities, under the same experimental setup.
Note that this scaling study is run on a different model generation from \qwenthreesixathreeb{}, so it is meant to complement rather than extends Table~\ref{tab:headline}.

\begin{table}[ht]
  \centering \small
  \setlength{\tabcolsep}{4pt}
  \begin{tabular}{l r ccccc}
    \toprule
    Model & Act. & Exec. & Search & Adapt. & Ambig. & Avg. \\
    \midrule
    \model{35B-A3B}    & 3B  & 10.1 & 22.0 &  5.3 &  4.4 & 10.5 \\
    \model{122B-A10B}  & 10B &  9.3 & 30.1 &  8.7 &  3.5 & 12.9 \\
    \model{27B}        & 27B & \textbf{21.4} & 30.6 & 12.7 & \textbf{7.6} & 18.1 \\
    \model{397B-A17B}  & 17B & 20.7 & \textbf{34.3} & \textbf{17.4} & 7.5 & \textbf{20.0} \\
    \bottomrule
  \end{tabular}
  \caption{\qwenthreefive{} scale ladder: \passk{1} (\%) pooled over the eleven
  languages, per capability and averaged (\texttt{time} excluded), with the
  active-parameter count per size. Rows are ordered by average score; best per
  column in \textbf{bold}. Same protocol as the headline (\passk{1},
  \texttt{thinking\_effort}$=$low, operational judge J3); each cell uses a
  denominator of $160$ scenarios per language-capability pair. Per-language cells
  are in Appendix~\ref{app:qwen_ladder}, Table~\ref{tab:qwen_ladder_lang}.}
  \label{tab:qwen_ladder}
\end{table}

\paragraph{Scaling is monotonic, but a dense mid-size rivals the largest MoE.}
Average \passk{1} rises monotonically with the ladder: $10.5 \to 12.9 \to 18.1 \to 20.0$. 
\model{397B-A17B} is the strongest size on average and on search and adaptability. 
But the dense \model{27B} actually edges the $6\times$-larger \model{397B-A17B} on execution ($21.4$ vs $20.7$) and ambiguity ($7.6$ vs $7.5$), and clears \model{122B-A10B} comfortably. 
Aggregate capability thus tracks the \emph{active} parameter count, and the dense mid-size model is fully competitive with the largest sparse one per active parameter.

\paragraph{The capability ordering is stable.} The canonical ordering of search $>$ execution $>$ adaptability $>$ ambiguity holds across sizes, with ambiguity hardest throughout; 
the one exception is \model{122B-A10B}, whose execution collapses to the level
of its adaptability. 
Scale lifts the whole profile rather than re-ordering which capabilities are easy.

\paragraph{The cross-lingual gap does not close with scale.} English is the ceiling at every size, and the gap to the target-language mean \emph{widens} as models grow: English minus the mean of the other ten languages is $+8.1$, $+7.5$, $+10.4$, and $+12.7$\,pp for \model{35B-A3B}$\to$\model{397B-A17B}.
\lang{hin} and \lang{jpn} are the weakest languages at every size (e.g.\ \model{397B-A17B}: \lang{eng} $31.6$ vs \lang{hin} $9.2$, \lang{jpn}
$10.0$), mirroring the script-family fragility of \S\ref{sec:analysis-capability}. 
Scaling the model therefore raises the multilingual floor but leaves the \emph{shape} of the cross-lingual gap intact, so the gap identified in this work is not one that a larger same-family checkpoint closes on its own.

\section{Error Attribution Gap: automatic and linguistic analysis}
\label{sec:attributionalgap}

The leaderboard numbers (\S\ref{sec:results}) surface a cross-lingual gap; this section asks \emph{why} it exists. First, we define an error taxonomy; second, we do an automatic evaluation, and finally, we perform human linguistic analysis.

\subsection{Error Taxonomy} 
\label{sec:taxonomy} 
To attribute each multilingual regression to a root cause, we apply a five-category verdict taxonomy, summarised below. 
Categories are checked top-down; the first that fits is assigned. 

\begin{enumerate} 

\item \textbf{Infrastructure (I1--I2)} --- The run crashed or no judgement was persisted; no gradable attempt exists.

\item \textbf{Translation defect (T1--T4)} --- The MT pipeline corrupted the task inputs (task description, universe app state field, or oracle event) such that the target run was set up to fail. Sub-codes cover oracle-event mistranslation~(T1), universe app-state corruption~(T2), task description ambiguity or meaning change~(T3), and cross-reference/transliteration breakage~(T4).\footnote{In practice T4 is near-empty as a distinct \emph{construction-level} defect: a dedicated non-Latin script audit finds no transliteration errors introduced by benchmark construction (App.~\ref{script_analysis}), and the few cross-reference cases we observe share the T1 mechanism---an oracle-compared literal left translated rather than passed through unchanged.}
\item \textbf{Verifier artefact (K1--K3)} --- The agent's action was effectively correct but the verifier rejected it---typically a translation-sensitive \emph{judge} (LLM-as-Judge) rejection (K1), a locale/format mismatch on a hard checker (K2), or list/ordering strictness (K3).
\item \textbf{Agent failure (A1--A7)} --- The model genuinely erred on a fair, correctly-translated task. 
It includes: wrong content/args~(A1), missing required action~(A2), extra/spurious action~(A3), wrong tool/target~(A4), wrong final answer~(A5), refusal/no answer~(A6), non-convergence~(A7). 
\item \textbf{Inconclusive (E1)} --- Multiple plausible causes coexist or the deciding evidence (e.g.\ the judge rationale) is unavailable.

\end{enumerate} 

\subsection{Automatic estimation}
\label{sec:automaticgap}

To scale error attribution beyond hand-picked cases, we automatically triage every cross-lingual failure into the taxonomy of \S\ref{sec:taxonomy} and estimate how the cross-lingual gap divides across its three fault sides. 
After correcting for infrastructure aborts, stratifying regressions by within-scenario determinism, and extrapolating to scenarios that fail in \emph{both} languages, we find the gap to be majority genuine model failure ($55.4\%$), with translation defects a smaller and spatially localised contamination floor ($34.5\%$ of the gap; only $6.4\%$ of all benchmark pairs rendered unsolvable in the target) and verifier artefacts the remainder ($10.1\%$). 
Five-annotator human re-adjudication confirms the automatic fault-side label on $91.4\%$ of a stratified sample. 
The remainder of this subsection develops each step.

\paragraph{Triage protocol.}
We attribute each target-language failure to the taxonomy of Section~\ref{sec:taxonomy} with an LLM-agent triage protocol that operates on the per-scenario artefacts persisted by the evaluation harness: the graded verdict, the environment action log (the agent's write actions), the agent trajectory, and, where available, the LLM-as-Judge rationale together with the oracle reference. 
One diagnostic agent handles one unit of work: it reads these artefacts and, in \emph{regression mode}, diffs a failing target run against a passing English run for the same scenario---the English run is the ground truth the dataset does not otherwise provide---before assigning a category, sub-code, and confidence. 
Judge rejections, which are otherwise ambiguous between a verifier artefact (K1) and a wrong final answer from the agent (A5), are adjudicated from the persisted judge rationale and the oracle reference rather than inferred from which check fired.
The protocol is orchestrated as an extract--diagnose--compile fan-out and is described in full in Appendix~\ref{app:triage-skill}.

\paragraph{Gap-representative stratification.}
The severity of a regression is itself informative. 
A deterministic translation defect corrupts the task and breaks \emph{every} target run, whereas a stochastic model slip or a flaky judge breaks only \emph{some}.
Sampling only complete collapses (English 3/3, target 0/3) therefore over-represents translation defects. 
We instead stratify infra-clean regressions by within-scenario determinism---\emph{deterministic} (the target never succeeds) versus \emph{stochastic} (the target succeeds on a strict subset of runs)---triage a language-\ and capability-balanced sample of each at the granularity of individual failing runs, and reweight each stratum's cause composition by its true failing-run volume (known exactly from the run index) to obtain a gap representative estimate. Confidence intervals are obtained by a scenario-clustered bootstrap.

\paragraph{Composition of the gap.}
The two strata differ sharply (Table~\ref{tab:strata}): deterministic failures are translation-driven ($59\%$), but stochastic failures---which carry the larger share of the gap---are model-driven ($75\%$).
After reweighting (Table~\ref{tab:auto-split}), \textbf{$55.4\%$} (95\% CI $50.3$--$60.7$) of the cross-lingual gap is a genuine model failure, \textbf{$34.5\%$} ($29.5$--$39.1$) a translation defect, and \textbf{$10.1\%$} ($7.1$--$13.5$) a verifier artefact. Restricting attention to complete collapses inverts this to a translation-dominated $\approx\!55\%/33\%$ split, quantifying the selection bias that the stratification removes. The model-driven share is language-dependent, largest for Turkish and Japanese and smallest for Portuguese and Spanish
(Appendix~\ref{app:triage-skill}).

\begin{table}[h]
\centering\small
\begin{tabular}{lccc}
\toprule
\textbf{Stratum} & \textbf{Translation} & \textbf{Verifier} & \textbf{Agent} \\
\midrule
Deterministic (target never solves) & 59\% & 8\%  & 33\% \\
Stochastic (target solves some runs) & 12\% & 13\% & 75\% \\
\bottomrule
\end{tabular}
\caption{Cause composition of genuine target failures by regression stratum.
Deterministic collapses are translation-driven; stochastic partial regressions are model-driven. 
Sampling only the former biases the estimate towards translation.}
\label{tab:strata}
\end{table}

\begin{table}[h]
\centering\small
\begin{tabular}{lcc}
\toprule
\textbf{Fault side} & \textbf{Share of gap} & \textbf{95\% CI} \\
\midrule
Agent failure (model) & \textbf{55.4\%} & [50.3, 60.7] \\
Translation defect    & 34.5\%          & [29.5, 39.1] \\
Verifier artefact     & 10.1\%          & [7.1, 13.5]  \\
\bottomrule
\end{tabular}
\caption{Reweighted, infra-clean decomposition of the cross-lingual gap
(Claude~4.7, ten target languages, all capabilities). Infrastructure failures
($8.5\%$ of runs) are excluded; CIs from a scenario-clustered bootstrap.}
\label{tab:auto-split}
\end{table}

\paragraph{From regressions to the whole benchmark.}
The triage above conditions on English \emph{passing}; by construction it cannot see scenarios that fail in \emph{both} languages, which is precisely where a translation defect could deflate a target score without leaving a visible
regression. 
To bound contamination over the entire benchmark we therefore partition all $6{,}056$ English$\times$target scenario--language pairs by outcome
(Appendix \ref{app:triage-skill} Table~\ref{tab:contingency}) and use a simple observation: a translation defect can only corrupt a score by rendering the target \emph{unsolvable}, i.e., the target never passes. 
The $74.5\%$ of pairs solved in the target on at least one run are thus translation-clean by construction---the
translated task is demonstrably solvable---leaving only the $25.5\%$ the target never solves to account for.

\paragraph{The both-fail blind spot.}
We triage all three ``target-never-solves'' cells, including the both-fail cell that regression analysis misses. 
The translation-defect rate falls monotonically as English competence drops: $59\%$ where English solves the scenario (a clean regression), $16\%$ where English is already flaky, and only $10\%$ (95\% CI $5.6$--$16.9$) where English also fails. In other words, translation defects concentrate exactly where they are already \emph{visible} as clean regressions, and the blind spot is if anything \emph{cleaner} than the part we can see---ruling out a hidden reservoir of contamination---and in any case both-fail defects cannot bias the cross-lingual ranking, since a scenario English also fails contributes nothing to the gap.
Weighting the three cells by size, translation defects render only \textbf{$6.4\%$} of all pairs unsolvable in the target, clustered on a handful of oracle- and keyword-mistranslation bugs. 
The cross-lingual gap therefore predominantly reflects genuine model capability, over a small, localised, and fixable translation-contamination floor. 

\paragraph{Human-based validation.}
To test the triage beyond hand-picked cases, five annotators re-adjudicated a stratified sample of triaged failures through a purpose-built review interface (Appendix~\ref{app:triage-validation}), each confirming or rejecting the assigned fault side from decision-relevant evidence alone (source/translated task, oracle, judge rationale, agent answer). 
Across \textbf{140} items in seven languages, they confirmed the automatic label in \textbf{128}---\textbf{91.4\%} agreement (95\% CI $85.6$--$95.0$)---with comparable rates across all three sides  (Translation $19/22$, Verifier $25/27$, Agent $84/91$) and rejections spread across category
boundaries with no systematic direction. 
This corroborates the decomposition of Table \ref{tab:auto-split}: independent human judgement agrees with the automatic
fault-side assignment on the large majority of a representative sample; full breakdown in Appendix~\ref{app:triage-validation}.

\subsection{Linguistic Analysis}
\label{sec:linguisticanalysis}
  Three linguists, native in Indonesian, Mandarin Chinese, and Spanish respectively—all proficient in English, with the Mandarin speaker also proficient in Japanese---independently inspected agent traces, translated prompts, and universe content for 
  16 scenarios spanning all four capabilities; six representative cases are presented in Table~\ref{tab:merged-cases}.

\begin{table*}[ht!] \centering \small \caption{Cross-lingual regression cases with per-language failure analysis (\claude{}, \passk{3}). Bold marks regressions relative to English.} \label{tab:merged-cases} \begin{tabular}{clp{2.4cm}ccccccp{5.2cm}} \toprule \textbf{\#} & \textbf{Cap.} & \textbf{English prompt} & \textbf{eng} & \textbf{spa} & \textbf{cmn} & \textbf{jpn} & \textbf{ind} & \textbf{Code} & \textbf{Per-language failure} \\ \midrule 1 & search & \emph{``events whose \textbf{titles} begin with `Research'\,''} & 3/3 & \textbf{0/3} & \textbf{0/3} & \textbf{0/3} & 3/3 & T3+T2 & \textbf{spa/cmn/jpn}: ``titles'' rendered as job/position (\emph{cargos}/\begin{CJK}{UTF8}{gbsn}职位/役\end{CJK}\begin{CJK}{UTF8}{bsmi}職)\end{CJK};%
agent filters by job title $\to$ empty set. \textbf{ind}: same T3 (\emph{gelar}$\neq$\emph{judul}) but agent reads event-title sense; passes via reinterpretation despite T2 word-order break on prefix match. \\[6pt] 2 & search & \emph{``cities I booked a cab \textbf{to} this month''} & 1/3 & \textbf{1/3} & \textbf{0/3} & \textbf{0/3} & \textbf{0/3} & T3+I1 & \textbf{cmn/jpn/ind}: directional ``to'' dropped outright; includes out-of-scope pickup city. \textbf{spa}: ``a las que\,...\,reservar'' retains surface form but \emph{reservar} blocks motion reading $\to$ agent flags ambiguity, still fails. \textbf{eng}: 1 pass + 2 infra aborts (1/1 on completed). \\[6pt] 3 & exec. & \emph{``Save 13 apartments, average price, email broker''} & 3/3 & \textbf{1/3} & 3/3 & \textbf{1/3} & 2/3 & K1+A3+A2 & \textbf{jpn}: judge rejects omitted currency unit in 2/3 runs (K1, stochastic; 3rd run omits same unit and passes). \textbf{ind}: correct end state but count-gate fails 14-vs-13 idempotent dup (A3). \textbf{spa}: agent skips save step entirely despite mentioning it in reasoning (A2, genuine). \\[6pt] 4 & adapt. & \emph{``Email updated guest list''} & 3/3 & 2/3 & 3/3 & 2/3 & 2/3 & K1+A3 & \textbf{spa}: judge hallucinates misspelling ``Dupon'' never produced; agent output correct ``Dupont'' (K1, fabricated rejection). \textbf{jpn/ind}: self-correction (omit $\to$ redo) reaches correct state but superseded write trips tool-count gate (A3). \\[6pt] 5 & ambig. & \emph{``Save condos; email contacts who wrote `Dobry den' in August''} & 2/3 & \textbf{0/3} & \textbf{0/3} & 1/3 & 1/3 & A3+T2 & \textbf{All langs}: ambiguity correctly handled (agents ask for clarification). \textbf{spa/jpn/ind (+eng 1/3)}: count-gate 6-vs-1 on redundant saves (A3); not cross-lingual. \textbf{cmn}: condo property-type enum inconsistently labelled \begin{CJK}{UTF8}{gbsn}(公寓 vs 共管公寓)\end{CJK}%
; filter unreachable (T2) + infra abort. \\[6pt] 6 & ambig. & \emph{``Delete \textbf{my contact} from the US''} (2 US contacts) & 3/3 & 2/3 & \textbf{0/3} & \textbf{0/3} & \textbf{0/3} & T3 & \textbf{eng/spa}: singular+article preserves one-vs-many cue; agent detects mismatch, \emph{asks} (deletes 0). \textbf{cmn/jpn/ind}: no article / no obligatory plural $\to$ ``my contact'' reads as generic set; agent perceives no conflict, \textbf{deletes both} (irreversible over-action). \textbf{spa} (1 failure): agent retrieval miss (finds only 1 contact), not MT. \\ \bottomrule \end{tabular} \end{table*}

We organise the observations by the three non-infrastructure fault sides of the taxonomy (\S\ref{sec:taxonomy}), mirroring the automatic decomposition of \S\ref{sec:automaticgap}.

\paragraph{Translation defects (T).} \textbf{Translation neutralises disambiguating surface cues.} Several prompts are ambiguous in English yet resolvable through word order or morphology; translation collapses these cues and forces a single reading that may not be the intended one (``whose titles'' rendered as \emph{job title} rather than \emph{event title} in every target language; Case~1, T3). \textbf{Faithful-looking translations silently drop constraints.} A directional preposition (``booked a cab \emph{to}'') was dropped or weakened in every target language, removing the very constraint that fixes the correct answer (Case~2, T3); the Spanish ``a las que\ldots reservar'' \emph{looks} directional but is incongruent with the verb's non-motion reading. \textbf{Morphological cue loss drives the most consequential failures.} In article-less, number-neutral languages (cmn/jpn/ind) the singular ``my contact'' is parsed as a generic set, and the agent deletes \emph{both} matching contacts instead of asking for clarification (Case~6, T3).%
Controlled-vocabulary drift compounds this: an inconsistently rendered property-type enum leaves the target filter unreachable (Case~5, T2).

\paragraph{Verifier artefacts (K).} \textbf{Judge strictness is language-dependent, stochastic, and can hallucinate.} The soft judge rejects correct target-language content while accepting identical English content, and in one case cited agent text that was never produced (Case~4, K1). \textbf{Action-count gating penalises harmless self-corrections.} The hard count-gate fails any mismatch between the agent's and the oracle's write-action counts even when the final state is correct---e.g.\ an agent that omits an attendee, deletes the event, and recreates it correctly (Cases~3--5). Although the taxonomy records these as extra-action agent errors (A3), the correct end state means they are largely \emph{measurement} artefacts of end-state-blind grading rather than genuine capability regressions.

\paragraph{Agent failures (A).} \textbf{Genuine model errors are present and, in aggregate, dominant.} Consistent with the automatic decomposition (\S\ref{sec:automaticgap}), where model failures account for $55.4\%$ of the cross-lingual gap, the sample contains clean slips traceable to neither MT nor the verifier: a Spanish run skips a required save step despite naming it in its own reasoning (Case~3, A2), and a retrieval miss returns only one of two matching contacts (Case~6, spa). \textbf{Genuine over-actions must be separated from grading artefacts.} Not every extra write is benign: an over-action that changes the final state is a true A3 regression, unlike the idempotent self-corrections penalised by the count-gate above, and the two must be disentangled before an action-count failure is read as a capability gap.

Overall, the linguistic analysis illustrates each non-infrastructure fault side of \S\ref{sec:taxonomy} with concrete cross-lingual cases and is consistent with the automatic decomposition of \S\ref{sec:automaticgap}: translation defects and verifier artefacts are real but bounded, while genuine, translation-independent agent errors
surface even in this small qualitative sample.

\section{Conclusion}
\label{sec:conclusion}

We deliver \datasetname, a machine-translated multilingual expansion of the \gaiatwo agentic benchmark spanning ten priority target languages, together with a localised and human-calibrated verifier and a cross-lingual leaderboard for a heterogeneous seven-agent cohort spanning frontier closed-source systems and open-weights models of both dense and mixture-of-experts architectures.

Our main findings are as follows. 
First, the cross-lingual gap is universal in direction but agent-specific in magnitude (8.8–18.4 pp in \passk{3}), and no agent is uniformly robust: the frontier systems narrow the gap without closing it. 
The metric inspected differentiates them: \gemini{} is the most self-consistent (smallest \passallk{3} gap) while \claude{} keeps high any-of-three accuracy off English yet loses per-attempt reliability. 
Second, the gap is predominantly model-driven: a stratified error attribution assigns 55\% of it to genuine agent failures, 35\% to translation defects and 10\% to verifier artefacts, while a benchmark-wide bound leaves only 6.4\% of scenario–language pairs MT-unsolvable; the gap concentrates on tool-orchestration and response quality rather than quantitative or categorical reasoning. 
Third, this gap does not close with model scale: a same-family size ladder shows the English–target delta widening from +8 to +13 pp as active parameters grow. 
Fourth, agents adopt a more hesitant execution strategy off English, foraging longer over the environment and committing later and less decisively (the write-action share of tool calls falls by 1--2.5 pp); this shift is the behavioural signature of degraded comprehension: it accompanies the mis-counted write actions behind the execution-side failures rather than reflecting a successful adaptation.
Finally, linguistic analysis identifies morphological cue loss and amplified ambiguity as primary failure mechanisms, particularly in non-Latin-script languages.

Taken together, our results argue that multilingual agentic
evaluation belongs alongside multilingual understanding evaluation as a standard part of the reporting protocol for globally deployed agents.
Several directions remain open. The most immediate is to broaden coverage along two axes: capabilities (e.g. the one we deferred in this work) and target languages, ideally with a focus on the lower-resource end where we posit the cross-lingual gap is widest. 
A complementary direction is to harden the translation pipeline (\S\ref{sec:pipeline}) against the translation-defect mechanisms our error analysis surfaces (\S\ref{sec:linguisticanalysis}), driving down the $6.4\%$ translation-induced unsolvability floor of the benchmark.

\section*{Acknowledgements}
\label{sec:acks}

We thank David Dale for his valuable feedback in early versions of the paper.

\section*{Limitations}
\label{sec:limitations}

\begin{itemize}
  \item \textbf{Excluded capabilities.} 
  The three \gaiatwo capabilities we do not evaluate (Time, Noise and Agent2Agent) are initially out of scope for \datasetname by design (\S\ref{sec:setup}): their scenarios either lack a scripted oracle trajectory or require a distinct verifier path that our translation pipeline does not currently cover.
  Extending the pipeline and verifier localisation to those splits is future work.
  \item \textbf{Translation quality.} \datasetname is constructed by an automatic machine-translation pipeline in a translator-only configuration, without human post-editing of the full corpus. The on-pipeline signals of
  \S\ref{sec:pipeline} attest to translated-content quality but do not substitute for a thorough per-language MT-quality audit against professional human references. As a result, a bounded but non-zero residual translation-defect floor persists in the evaluated release (\S\ref{sec:automaticgap}), concentrated in non-Latin, morphologically rich target languages; although we show this floor to be relatively small, scores in the most affected languages should be read as carrying it.
  \item \textbf{Judge reliability ceiling.} Our judge is calibrated against human annotations, but human annotators themselves agree only moderately on this task. The judge's high agreement scores should therefore be read as parity with an imperfect human reference rather than as absolute correctness; a higher-agreement re-annotation of the calibration set would be needed to tighten this ceiling.
  \item \textbf{Closed-system extended thinking.} Our headline numbers evaluate the three proprietary systems we benchmark (\claude, \gptfive, \gemini) at the provider-default reasoning-effort setting, for budget reasons. 
  Our controlled ladder (\S\ref{sec:setup}) indicates that reasoning budget does not drive the multilingual gap, but it covers only one model and three target languages; extending the ladder to more languages is needed before drawing grounded conclusions about the role of reasoning effort. 
  We reserve this for future work.
  \item \textbf{Harness choice.} 
  All numbers are reported under the \textsc{OpenClaw} harness, with the exception of Kimi. These numbers should be read as observational comparisons under a fixed harness rather than as definitive model rankings; a different harness (e.g. co-developed with the model) could reorder the leaderboard.
  \item \textbf{Human audit.} A per-language human audit of cross-lingual regressions is available only on a few languages and samples (\S\ref{sec:attributionalgap}). A full per-language
  regression audit across the ten-language matrix is future work.
\end{itemize}

\section*{Ethics Statement}

\datasetname is an evaluation benchmark built on top of \gaiatwo and contains no personal data beyond the synthetic content of the original scenarios. 
Our goal is to improve the equity of AI agents by making cross-lingual performance gaps measurable and visible; documenting that agents underperform for non-English users is a prerequisite for closing that gap. We caution that machine-translated benchmarks can embed translationese and cultural mismatches, and we therefore do not treat \datasetname scores as a measure of culturally appropriate behaviour; this remains a concern for future research. 
We will release the dataset and human annotations to support reproducibility and further research.

\bibliographystyle{meta/assets/plainnat}
\bibliography{bibliography/custom}

\clearpage
\appendix
\section{Languages}
\label{app:languages}

Table \ref{tab:language_comparison} specifies the languages coverage of related multilingual agentic datasets, including ours.

\begin{table}[ht] 
\centering \scriptsize 
\begin{tabular}{lccc} 
\toprule 
\textbf{Language} & \textbf{MAPS} & \textbf{GAIA-v2-LILT} & \textbf{\datasetname} \\ 
\midrule 
Arabic & \checkmark & \checkmark & -- \\ 
German & \checkmark & \checkmark & \checkmark \\ 
Hindi & \checkmark & \checkmark & \checkmark \\ 
Korean & \checkmark & \checkmark & -- \\ 
Portuguese & \checkmark & \checkmark & \checkmark \\ 
Italian & \checkmark & -- & \checkmark \\ 
Japanese & \checkmark & -- & \checkmark \\ 
Spanish & \checkmark & -- & \checkmark \\ 
Russian & \checkmark & -- & -- \\ 
Hebrew & \checkmark & -- & -- \\ 
Mandarin Chinese & -- & -- & \checkmark \\ 
French & -- & -- & \checkmark \\ 
Indonesian & -- & -- & \checkmark \\ 
Turkish & -- & -- & \checkmark \\ 
\bottomrule \end{tabular} 
\caption{Language coverage comparison between MAPS \cite{hofman-etal-2026-maps}, GAIA-v2-LILT \cite{kim2026gaiav2liltmultilingualadaptationagent}, and our benchmark.} \label{tab:language_comparison} \end{table}
\section{The translation pipeline}
\label{app:translator}

In this section we report the results of the empirical studies we carry out that motivate the design of the final translation pipeline we adopt to produce the benchmark, as described in \S\ref{sec:pipeline}.
In particular, these motivate the choice of \gemma{} as the translator, and the choice to run it in a single translator-only pass with no downstream reviewer. Both decisions rest on the same pairwise protocol.
Every translatable span produced by a candidate configuration is compared against the corresponding span of a \emph{reference translation}, a fixed, high-quality rendering of the same content. The comparison is evaluated by an LLM judge that is tasked to return one of \{\emph{equivalent}, \emph{candidate-wins}, \emph{reference-wins}, \emph{both-wrong}\}.
In particular, we consider a fixed sample of $200$ spans, stratified by app, and score each (translator, language) combination against two LLM judges, \gptfive{} and \gemini{}; each judge's verdict is the majority over three position-randomised trials (thus controlling for the known first-position bias of pairwise LLM judges), so the two judges together contribute $400$ pooled verdicts per combination.
We report \emph{candidate-win} rate (judge prefers the candidate over the reference, higher is better) and \emph{reference-win} rate (lower is better).

Both candidate translators run under a $16$K-token context budget, within which each per-scenario prompt must accommodate the source spans, the generated translation, and the term table of \S\ref{sec:pipeline}. Since the raw term table---up to roughly a thousand entries per universe---would overflow this budget on its own, we cap it at its $200$ shortest-source entries; a reproducibility check on \lang{spa} finds that the capped and uncapped tables yield \gemma{} translations differing by under $2.5$\,pp on every judged cell, confirming the cap is quality-neutral. Within the prompt this table is rendered inline as the \texttt{Terminology} block shown in \Cref{fig:translation-prompt}, and is omitted for scenarios that share no terms across surfaces. Both translators are executed with the reasoning (``thinking'') mode disabled: left enabled, it consistently exhausts the $16$K-token budget on internal reasoning traces before emitting any parse-able translation.

\begin{figure}[t]
\centering
\begin{tcolorbox}
\begin{verbatim}
Translate the following text from {src_lang} to {tgt_lang}.

### Terminology (use these exact translations for the following terms):
- "{source span 1}" -> "{target rendering 1}"
- "{source span 2}" -> "{target rendering 2}"
  ...

### Text to translate:
{src_text}

Keep any proper nouns, person names, IDs, email addresses, URLs, file names, 
and technical identifiers unchanged.
When the text contains quoted references to app content 
(conversation titles, email subjects, event titles), 
use the exact translations from the terminology section above if provided.

Respond with ONLY the translated text, no explanations or comments.
\end{verbatim}
\end{tcolorbox}
\caption{The prompt supplied to the translator for every text span.
\texttt{\{src\_lang\}} and \texttt{\{tgt\_lang\}} are the source and target
language and \texttt{\{src\_text\}} is the span to translate; the
\texttt{Terminology} block is the per-scenario term table (\S\ref{sec:pipeline})
rendered inline (placeholder entries shown; the pipeline joins each pair with a
``$\rightarrow$'' arrow), present only when the scenario shares terms across
surfaces. Oracle tool-call arguments are translated with a structurally
identical variant that additionally conditions on the user task and returns a
JSON object keyed by argument index.}
\label{fig:translation-prompt}
\end{figure}

\subsection{Translator choice}
\label{app:translator-hth}

\begin{figure}[t]
    \centering
    \includegraphics[width=0.98\columnwidth]{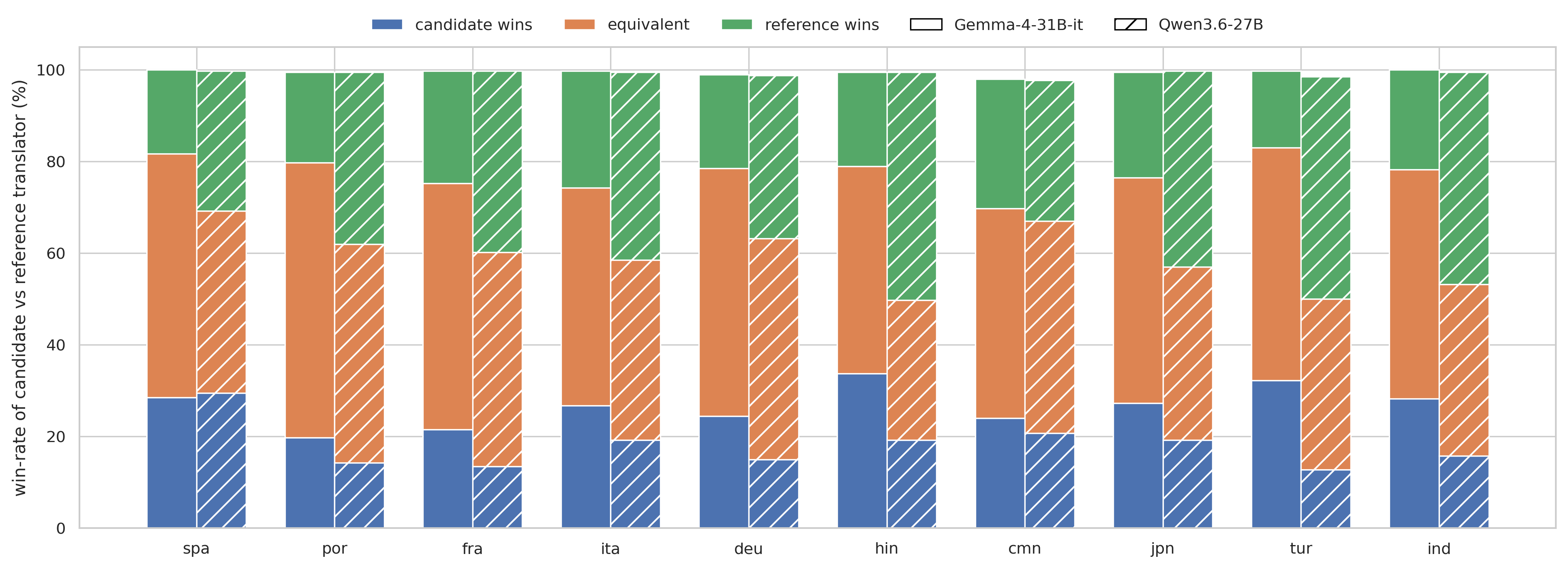}
    \caption{Translator head-to-head. Pooled $400$-verdict breakdown for
    \gemma{} (solid bars) and \qwenthreesixdense{} (hatched bars) on each of
    the ten target languages, each translation judged pairwise against the
    reference translation. The reference-wins share (top segment) is smaller
    for \gemma{} on every language, and the gap widens on the non-Latin and
    morphologically heavy targets.}
    \label{fig:translator-hth}
\end{figure}

We evaluate the two top-ranking open-weight systems on BOUQuET for our target languages: \gemma{} and \qwenthreesixdense{}, scoring translation across all ten languages from both systems against the reference translations under the protocol above.
\Cref{tab:translator-hth} reports the per-language verdict shares and \Cref{fig:translator-hth} visualises them.

We find that \gemma{} dominates \qwenthreesixdense{} on every one of the ten target languages: its reference-win rate is lower and its candidate-win rate higher on all ten. 
The margin is narrowest on the closest-to-saturation Latin-script targets (a $-6$\,pp reference-win gap on \lang{spa}) and widens sharply on the non-Latin and morphologically heavy targets ($-32$\,pp on \lang{tur}, $-29$\,pp on \lang{hin}), where surface-fidelity errors are harder to recover from. \gemma{} consistently wins on all ten languages and therefore we adopt it as our main and only translation system.

\begin{table}[t]
    \centering
    \small
    \begin{tabular}{lcccccc}
    \toprule
    & \multicolumn{3}{c}{\gemma} & \multicolumn{3}{c}{\qwenthreesixdense{}} \\
    \cmidrule(lr){2-4} \cmidrule(lr){5-7}
    Lang & cand & ref & equiv & cand & ref & equiv \\
    \midrule
    \lang{spa} & \textbf{28.5} & \textbf{18.2} & 53.2 & 29.5 & 30.5 & 39.8 \\
    \lang{por} & \textbf{19.8} & \textbf{19.8} & 60.0 & 14.2 & 37.5 & 47.8 \\
    \lang{fra} & \textbf{21.5} & \textbf{24.5} & 53.8 & 13.5 & 39.5 & 46.8 \\
    \lang{ita} & \textbf{26.8} & \textbf{25.5} & 47.5 & 19.2 & 41.0 & 39.2 \\
    \lang{deu} & \textbf{24.5} & \textbf{20.5} & 54.0 & 15.0 & 35.5 & 48.2 \\
    \lang{hin} & \textbf{33.8} & \textbf{20.5} & 45.2 & 19.2 & 49.8 & 30.5 \\
    \lang{cmn} & \textbf{24.0} & \textbf{28.2} & 45.8 & 20.8 & 30.8 & 46.2 \\
    \lang{jpn} & \textbf{27.2} & \textbf{23.0} & 49.2 & 19.2 & 42.8 & 37.8 \\
    \lang{tur} & \textbf{32.2} & \textbf{16.8} & 50.8 & 12.8 & 48.5 & 37.2 \\
    \lang{ind} & \textbf{28.2} & \textbf{21.8} & 50.0 & 15.8 & 46.2 & 37.5 \\
    \bottomrule
    \end{tabular}
    \caption{Translator head-to-head: pooled verdict shares (\%,
    $n=400$ verdicts per cell) for \gemma{} and \qwenthreesixdense{} against the
    reference translations, per target language. \gemma{} wins on every
    language; \textbf{bold} marks the winner's candidate-win (higher better) and
    reference-win (lower better) columns.}
    \label{tab:translator-hth}
\end{table}

\subsection{The reviewer step}
\label{app:translator-reviewer}

We further investigate whether a two-stage \emph{translate-then-review} pipeline, in which a second model flags and post-edits the translator's output improves \datasetname{} and find that it does not. 

Fixing the translator to \gemma{}, we consider a reviewer stage with three different configurations: \gptoss{} as a cross-family reviewer on all ten languages, \gemma{} itself as a self-review baseline on a three-language subset (\lang{spa}, \lang{cmn}, \lang{ind}), and \qwenthreesixdense{} as an alternate cross-family reviewer on the same subset.
For each configuration we measure the change in candidate-win rate against the reviewer-free \gemma{}-only baseline, $\Delta = \hat{p}_{\text{reviewed}} - \hat{p}_{\text{baseline}}$.

Every one of the sixteen reviewer configurations is statistically indistinguishable from the reviewer-free baseline, since all the $95\%$ confidence intervals on $\Delta$ 
cross zero (see \Cref{tab:reviewer-noop}). 
On the ten-language \gptoss{} sweep the smallest uncorrected $p$-value is $0.47$; pooling all ten cells gives a mean $\Delta = +1.0$\,pp with a $95\%$ CI of $[-1.0, +3.0]$\,pp. 
We evaluate other two choices of reviewer model on a reduced sample of languages, and find that a different choice of the reviewer model does not significantly alter the outcome.

Manual inspection of the edits confirms the statistical picture: the reviewer rewrites under $9\%$ of fields, with changes that are almost entirely stylistic (gender inclusion, casing, idiom normalisation) rather than error repairs. We therefore commit to a translation pipeline for \datasetname{} with a single translator-only pass, eliminating a second model pass that roughly doubles per-scenario latency for no measurable quality gain.

\begin{table}[t]
    \centering
    \small
    \begin{tabular}{llrcr}
    \toprule
    Lang & Reviewer & $\Delta$ pp & $95\%$ CI & $p$-value \\
    \midrule
    \lang{spa} & \gptoss{} & $+2.25$ & $[-4.1, +8.6]$ & $0.49$ \\
    \lang{por} & \gptoss{} & $+1.25$ & $[-4.3, +6.8]$ & $0.66$ \\
    \lang{fra} & \gptoss{} & $+1.50$ & $[-4.3, +7.3]$ & $0.61$ \\
    \lang{ita} & \gptoss{} & $+2.00$ & $[-4.2, +8.2]$ & $0.53$ \\
    \lang{deu} & \gptoss{} & $+1.00$ & $[-5.0, +7.0]$ & $0.74$ \\
    \lang{hin} & \gptoss{} & $+0.75$ & $[-5.8, +7.3]$ & $0.82$ \\
    \lang{cmn} & \gptoss{} & $-0.75$ & $[-6.6, +5.1]$ & $0.80$ \\
    \lang{jpn} & \gptoss{} & $+2.25$ & $[-4.0, +8.5]$ & $0.48$ \\
    \lang{tur} & \gptoss{} & $+2.25$ & $[-4.3, +8.8]$ & $0.50$ \\
    \lang{ind} & \gptoss{} & $-2.25$ & $[-8.4, +3.9]$ & $0.47$ \\
    \midrule
    \lang{spa} & \gemma{}-self        & $+1.25$ & $[-5.0, +7.5]$ & $0.70$ \\
    \lang{cmn} & \gemma{}-self        & $+1.75$ & $[-4.2, +7.7]$ & $0.57$ \\
    \lang{ind} & \gemma{}-self        & $-1.75$ & $[-7.9, +4.4]$ & $0.58$ \\
    \lang{spa} & \qwenthreesixdense{} & $-0.75$ & $[-7.0, +5.5]$ & $0.81$ \\
    \lang{cmn} & \qwenthreesixdense{} & $+0.50$ & $[-5.4, +6.4]$ & $0.87$ \\
    \lang{ind} & \qwenthreesixdense{} & $-1.50$ & $[-7.7, +4.7]$ & $0.63$ \\
    \bottomrule
    \end{tabular}
    \caption{Reviewer sweep: per-(language, reviewer) change in candidate-win rate $\Delta$ against the reviewer-free \gemma{}-only baseline. 
    Every configuration's confidence interval crosses zero: no reviewer configuration produces a statistically detectable change on any language.}
    \label{tab:reviewer-noop}
\end{table}

\subsection{Translator-family confound}
\label{app:family-match}

Because \gemma{} serves both as the model in the translation pipeline (\S\ref{sec:pipeline}) and as one of the evaluated agents---and \qwenthreesixdense{} is likewise evaluated---a natural concern is a \emph{family-match} confound: an agent might enjoy a home-field advantage on data translated by its own model family, inflating its apparent capability. 
We test this directly with a $2\times2$ ablation crossing agent $\in\{\gemma,\ \qwenthreesixdense\}$ with translator $\in\{\text{\gemma{}-MT},\ \text{\qwenthreesixdense{}-MT}\}$, where \qwenthreesixdense{}-MT re-translates the identical scenarios through the same pipeline. 
The design spans all ten target languages at $1{,}920$ rollouts per cell ($40$ cells, $262$K context), and we form the difference-in-differences
\begin{equation}
\mathrm{DiD}=\bigl[G_{\text{\gemma{}-MT}}-G_{\text{\qwenthreesixdense{}-MT}}\bigr]-\bigl[Q_{\text{\gemma{}-MT}}-Q_{\text{\qwenthreesixdense{}-MT}}\bigr],
\end{equation}
where $G$ and $Q$ are the \gemma{} and \qwenthreesixdense{} agent scores; $\mathrm{DiD}>0$ would indicate a \gemma{} home-field advantage.

The agent gap is essentially invariant to which family translated the evaluation data (Table~\ref{tab:family-match}): the pooled DiD is $+0.29$\,pp. Both agents lose comparably---$4.1$\,pp (\gemma) and $3.8$\,pp (\qwenthreesixdense)---when the data switches from \gemma{}-MT to \qwenthreesixdense{}-MT, a symmetric \emph{translator-quality} effect (\gemma{} is the stronger translator, App.~\ref{app:translator-hth}) rather than a family-of-origin advantage. 
To summarise across languages and to separate a genuine family-match effect from the overall difference in translator quality, we fit an ordinary least-squares (OLS) regression over the $40$ cell-level observations ($4$ cells $\times$ $10$ languages). 
Writing $y$ for a cell's completion-robust \passk{1} (the pass rate among graded rollouts, $\mathrm{pass}/(\mathrm{pass}+\mathrm{fail})$), we regress
\begin{equation}
y = \beta_0 + \beta_1\,A + \beta_2\,T + \beta_3\,(A\cdot T) + \varepsilon,
\end{equation}
where $A=1$ for the \gemma{} agent (and $0$ for \qwenthreesixdense{}) and $T=1$ when the evaluation data is \qwenthreesixdense{}-translated (and $0$ for \gemma{}-translated). 
The coefficients $\beta_1$ and $\beta_2$ are the two \emph{main effects}---the average change in \passk{1} from switching the agent, respectively the translator, with the other held fixed---and $\beta_3$, the coefficient on the product term, is the \emph{family-match interaction}: the extra change when agent and translator switch together, i.e.\ how much more the \gemma{} agent gains from \gemma{}-translated data than the \qwenthreesixdense{} agent does. 
Under this coding $\beta_3=-\mathrm{DiD}$, so a \gemma{} home-field advantage would show up as $\beta_3<0$.

\begin{figure}[h]
\centering
\includegraphics[width=0.98\columnwidth]{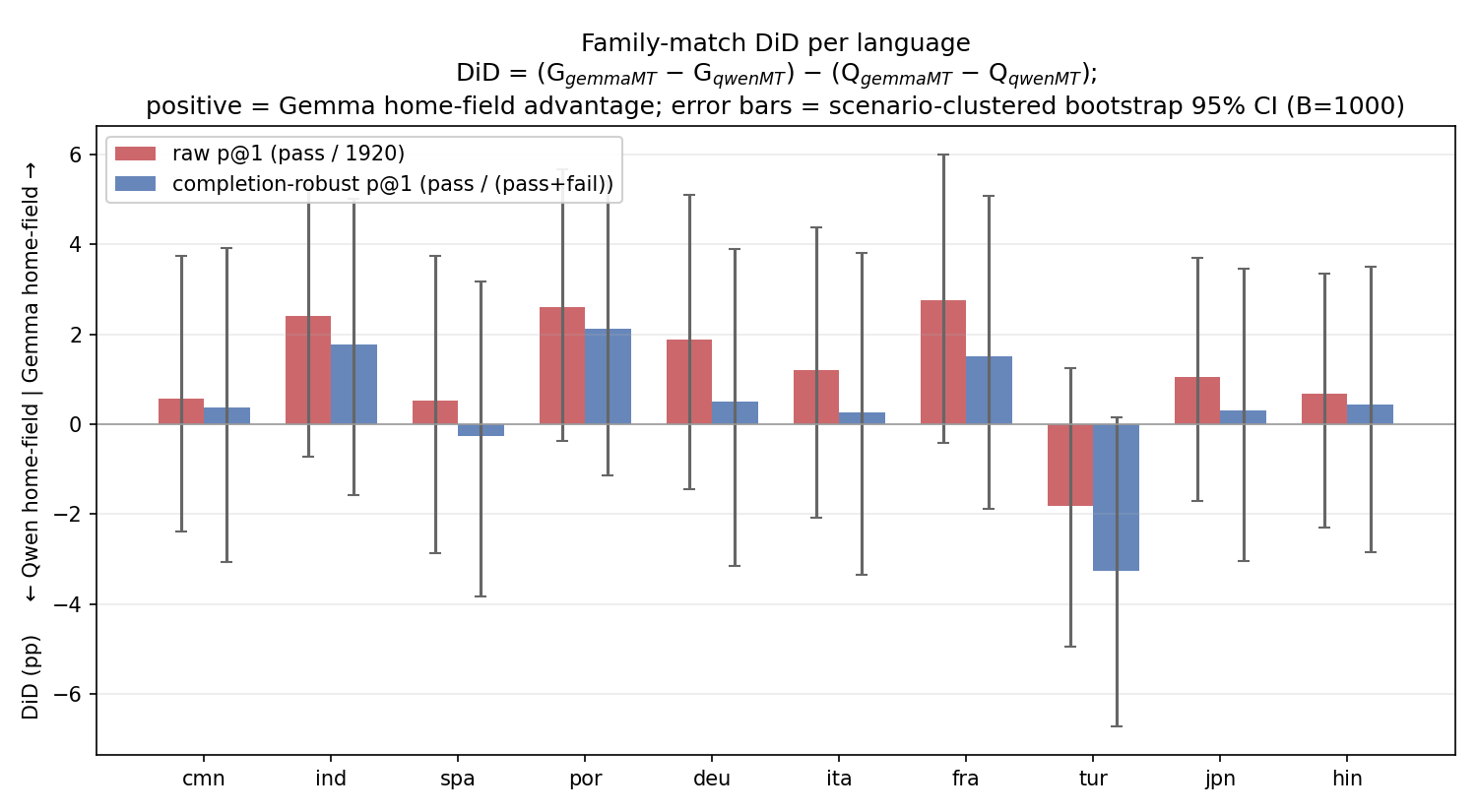}
\caption{Per-language difference-in-differences (\gemma{} home-field advantage), raw and completion-robust, with $95\%$ scenario-clustered bootstrap intervals ($B=1000$). All ten intervals cross zero under both metrics; positive values would indicate a \gemma{} family-match advantage. The one language with the largest magnitude (\lang{tur}) is Qwen-favoured.}
\label{fig:family-match}
\end{figure}

We estimate $\beta_3=-0.37$\,pp (implied $\mathrm{DiD}=+0.37$\,pp) with a standard error (SE, the estimated sampling uncertainty of the coefficient) of $2.9$\,pp; the corresponding $t$-statistic $t=\beta_3/\mathrm{SE}\approx0.13$ is far below the $|t|\!\approx\!2$ required for significance at the $5\%$ level, so the interaction is statistically indistinguishable from zero. Re-fitting with an added covariate---each cell's measured translator quality, taken as the per-language candidate-win rate of Table~\ref{tab:translator-hth}---leaves $\beta_3$ essentially unchanged, so the (already negligible) family-match signal is not an artefact of translator quality varying across languages. Complementing these pooled and regression estimates, every one of the ten \emph{per-language} DiD values has a $95\%$ confidence interval---from a scenario-clustered bootstrap (resampling whole scenarios with replacement, $B=1000$ times, so that runs of the same scenario stay together)---that crosses zero under both the completion-robust and the raw metric (Figure~\ref{fig:family-match}); the largest per-language effect (\lang{tur}, $-3.3$\,pp on the robust metric) is \emph{Qwen}-favoured, i.e.\ directionally opposite to a \gemma{} home-field advantage. We conclude the cross-lingual gap decomposition (\S\ref{sec:attributionalgap}) is not confounded by translator family.

\paragraph{Metric note.} The \gemma{}$\times$\qwenthreesixdense{}-MT cells carry ${\sim}13\%$ non-graded rollouts (agent timeouts and no-verdict cases on long search trajectories at $262$K context), asymmetrically concentrated in that single arm. We therefore take completion-robust \passk{1}${}=\text{pass}/(\text{pass}+\text{fail})$ as the primary metric and report raw \passk{1} as a robustness bound; the verdict is identical under both (pooled DiD $+0.29$ vs $+1.18$\,pp), and the difference-in-differences subtracts out most of the completion-rate asymmetry.

\begin{table}[h]
\centering\small
\begin{tabular}{lcc}
\toprule
& \multicolumn{2}{c}{\textbf{Evaluation data translated by}} \\
\cmidrule(lr){2-3}
\textbf{Agent} & \gemma{}-MT & \qwenthreesixdense{}-MT \\
\midrule
\gemma              & $31.2$ & $27.2$ \\
\qwenthreesixdense  & $32.7$ & $28.9$ \\
\bottomrule
\end{tabular}
\caption{Family-match $2\times2$: pooled completion-robust \passk{1} (\%) over all ten target languages ($1{,}920$ rollouts per cell). Both agents drop ${\sim}4$\,pp on \qwenthreesixdense{}-translated data; the difference-in-differences (\gemma{} home-field) is $+0.29$\,pp. Rows are the evaluated agent; columns are the translator that produced the evaluation data.}
\label{tab:family-match}
\end{table}

\clearpage
\clearpage

\section{Judge calibration}
\label{app:judge_calibration}

In this section we report additional details on the judge model and prompt ablation results described in \S\ref{sec:verifier-calibration}, on the original \gaiatwo human-annotated calibration set.

\subsection{English calibration}

The four configurations cross the two judge models, the paper's reference \llamareference and the new \gptoss, with the two prompt sets, the upstream default and our localised overrides:
J1 (\llamareference, default), J2 (\gptoss, default), J3 (\gptoss, localised), and J4 (\llamareference, localised).
Table~\ref{tab:calib_overall} reports overall Cohen's $\kappa$~\citep{cohen1960kappa} against the human-majority label on the $n=414$ non-inconclusive traces, with bootstrap $95\%$ confidence intervals.
J3 is the judge configuration we use when constructing the agent leaderboard in~\S\ref{sec:results}.

\begin{table}[h]
\centering\small\setlength{\tabcolsep}{3pt}
\begin{tabular}{lllcccc}
\toprule
\textbf{Config} & \textbf{Model} & \textbf{Prompts} & \textbf{Agreement} & $\kappa$ & $95\%$~CI & $\Delta$pass \\
\midrule
J1 & \llamareference & default & $0.879$ & $0.714$ & $[0.641, 0.781]$ & $-0.111$ \\
J2 & \gptoss         & default & $0.882$ & $0.720$ & $[0.644, 0.787]$ & $-0.109$ \\
J3 & \gptoss         & localised & $0.886$ & $0.732$ & $[0.661, 0.796]$ & $-0.104$ \\
J4 & \llamareference & localised & $0.882$ & $0.720$ & $[0.647, 0.789]$ & $-0.109$ \\
\bottomrule
\end{tabular}
\caption{Overall judge agreement against the human-majority label on the $n=414$ non-inconclusive traces of the recovered calibration set. All four $\kappa$ values fall in $[0.714, 0.732]$ with heavily overlapping CIs: on English, neither the model swap nor prompt localization materially changes
agreement with humans. J3 matches the paper-standard reference J1.}
\label{tab:calib_overall}
\end{table}

Because all four configurations are scored on the same frozen traces, the two design choices can be isolated directly by judge-vs-judge agreement.
Holding the model fixed at \llamareference, localizing the prompts (J1$\rightarrow$J4) flips the verdict on \emph{one single} trace out of $462$
($\kappa=0.994$, $95\%$~CI~$[0.978, 1.000]$); holding the prompts fixed at the localised set, swapping the model (J4$\rightarrow$J3) flips two ($\kappa=0.987$, $95\%$~CI~$[0.967, 1.000]$); and the full swap
from the paper's reference (J1$\rightarrow$J3) flips three ($\kappa=0.981$, $95\%$~CI~$[0.957, 1.000]$), every one in the direction of J3 accepting a trajectory the reference rejected. On English, then, both the model choice and the prompt version are immaterial to judge--human agreement; the localisation payoff is realised off-English (\S\ref{sec:verifier-calibration}).

Table~\ref{tab:calib_percap} reports per-capability $\kappa$ against the human-majority label alongside the human inter-annotator ceiling where multi-annotator coverage exists on this set.

\begin{table}[h]
\centering\small\setlength{\tabcolsep}{4pt}
\begin{tabular}{lccccc}
\toprule
\textbf{Capability} & Human IAA & J1 & J2 & J3 & J4 \\
\midrule
adaptability & $0.327$          & $0.000$ & $0.000$ & $0.000$ & $0.000$ \\
ambiguity    & (single-annotation)  & $0.766$ & $0.798$ & $0.798$ & $0.798$ \\
execution    & (single-annotation)  & $0.595$ & $0.619$ & $0.642$ & $0.595$ \\
search       & (single-annotation)  & $1.000$ & $0.979$ & $1.000$ & $1.000$ \\
time         & $0.104$          & $0.000$ & $0.000$ & $0.000$ & $0.000$ \\
\bottomrule
\end{tabular}
\caption{Per-capability Cohen's $\kappa$ against the human-majority label under each judge configuration. The Human-IAA column is Cohen's $\kappa$ on the multi-annotator subset of the calibration set (adaptability $n=83$, time $n=45$); the other three capabilities are single-annotator on this set,
so no human ceiling can be computed.}
\label{tab:calib_percap}
\end{table}

The $\kappa=0$ entries on \emph{adaptability} and \emph{time} appear in \emph{all four} configurations, not only under J4. On these capabilities the judge---like the human annotators-- passes essentially no trace, so the label distribution has near zero variance and Cohen's $\kappa$ collapses to $0$ regardless of raw agreement (which exceeds $0.87$ throughout). 
This is a property of the calibration corpus, not a judge weakness; see the failure-taxonomy discussion below.

Because the full-corpus $\kappa$ of Table~\ref{tab:calib_overall} is depressed
both by these degenerate capabilities and, more broadly, by the
deterministic-checker-dominated traces on which every configuration returns the
same structural verdict, we recompute the $2\times2$ on the \emph{LLM-touched}
slice---the $n=153$ traces whose verdict actually invokes an LLM
checker.\footnote{A trace is LLM-touched if it passes, or if its first failing
turn is decided by one of the natural-language checkers
(\texttt{message\_checker}, \texttt{user\_message\_checker},
\texttt{signature\_checker}, \texttt{tone\_checker}, \texttt{content\_checker})
or returns inconclusive; \texttt{time} is excluded, as in the multilingual
slice.} This is the \emph{same} frozen slice used for the multilingual
comparison (Table~\ref{tab:calib_multilingual_llm}), so the English and
cross-lingual numbers are directly comparable. On it
(Table~\ref{tab:calib_overall_llm}) every configuration rises to
$\kappa \geq 0.928$, against $0.714$--$0.732$ on the full corpus: the
operational judge is strongest precisely where the LLM is exercised, and the
full-corpus figures understate rather than overstate its reliability. The
$2\times2$ ordering is preserved---localised $\geq$ default on each model, and
\gptoss $\geq$ \llamareference under each prompt set---though all four CIs
overlap, so the differences are directional, not significant.

\begin{table}[h]
\centering\small\setlength{\tabcolsep}{3pt}
\begin{tabular}{lllccccc}
\toprule
\textbf{Config} & \textbf{Model} & \textbf{Prompts} & $n$ & Agr. & $\kappa$ & $95\%$~CI & $\Delta$pass \\
\midrule
J1 & \llamareference & default   & $153$ & $0.967$ & $0.928$ & $[0.859, 0.985]$ & $-0.007$ \\
J2 & \gptoss         & default   & $153$ & $0.974$ & $0.942$ & $[0.881, 0.986]$ & $\phantom{-}0.000$ \\
J3 & \gptoss         & localised & $153$ & $0.987$ & $0.971$ & $[0.923, 1.000]$ & $+0.013$ \\
J4 & \llamareference & localised & $153$ & $0.974$ & $0.942$ & $[0.881, 0.986]$ & $\phantom{-}0.000$ \\
\bottomrule
\end{tabular}
\caption{The $2\times2$ recomputed on the LLM-touched slice ($n=153$; the same
frozen traces as the multilingual Table~\ref{tab:calib_multilingual_llm}).
Removing the deterministic-checker-dominated and degenerate-$\kappa$
capabilities lifts every configuration to $\kappa \geq 0.928$, versus
$0.714$--$0.732$ on the full corpus (Table~\ref{tab:calib_overall}). Localising
the prompts on the fixed \llamareference model (J1$\rightarrow$J4) still moves
the verdict on essentially no trace ($\kappa=0.986$ judge-vs-judge on this
slice).}
\label{tab:calib_overall_llm}
\end{table}

\subsection{Failure taxonomy on the calibration set}

The calibration set is heavily deterministic-checker-dominated on three of the four capabilities: on adaptability, ambiguity and execution the verdict is decided by structured checks that never exercise the LLMaaJ component. Table~\ref{tab:failure_taxonomy} reports the failure distribution per capability.

\begin{table}[h]
\centering\small\setlength{\tabcolsep}{4pt}
\begin{tabular}{lccc}
\toprule
\textbf{Capability} & Success & Det.\ fail & LLM fail \\
\midrule
adaptability & $0$  & $80/83$ ($96\%$) & $3/83$ ($4\%$) \\
ambiguity    & $19$ & $72/95$ ($76\%$) & $4/95$ ($4\%$) \\
execution    & $29$ & $64/94$ ($68\%$) & $1/94$ ($1\%$) \\
search       & $54$ & $0$              & $43/97$ ($44\%$) \\
\bottomrule
\end{tabular}
\caption{Verdict provenance on the English calibration set, computed on the operational judge J3 (not the human labels). \emph{Success} counts traces J3 passes; \emph{Det.\ fail} counts J3 failures whose first failing check is deterministic (tool- or SMU-count, stuck-loop or timeout, or a deterministic content checker); \emph{LLM fail} counts J3 failures whose first failing check is an LLM checker (\texttt{message}, \texttt{user\_message}, \texttt{signature}, \texttt{tone}, or \texttt{content}). The three counts sum to $n$ per capability. Only \emph{search} exercises the LLMaaJ component meaningfully.}
\label{tab:failure_taxonomy}
\end{table}

\subsection{Multilingual robustness}

Table~\ref{tab:calib_multilingual_llm} restricts the calibration set to
the $n=153$ traces where the LLM judge actually fires and reports
per-language $\kappa$ under J3 for \lang{eng} and all ten target
languages. This is the headline number of \S\ref{sec:verifier-calibration}:
it is the only slice on which translation of the natural-language content
can plausibly move the verifier verdict. Because the English human-majority
label is held fixed across languages (a faithful translation should not
change the correct verdict), $\kappa$ here isolates whether translation
alone perturbs the judge.\footnote{In roughly $2\%$ of translated traces
($76$ of $3{,}690$) the \gemma translator regurgitated part of its own
system prompt into the oracle user message, concentrated on the
\emph{search} capability whose user messages are longest. Manual review of
the $72$ affected traces that fall in the LLM-touched slice found no verdict
divergence from English on any of them, so the calibration figures reported
here are unaffected.} The eight languages beyond \lang{eng}, \lang{jpn}
and \lang{spa} were scored on a freshly provisioned \gptoss endpoint with
identical \texttt{gpt-oss-120b} weights and \texttt{omnigaia\_v2} prompts,
differing only in serving host.

\begin{table}[h]
\centering\small\setlength{\tabcolsep}{4pt}
\begin{tabular}{lcccc}
\toprule
\textbf{Language} & $n$ & $\kappa$ & Judge pass & $\Delta\kappa$ vs \lang{eng} \\
\midrule
\lang{eng} & $153$ & $0.971$ & $0.667$ & --- \\
\lang{cmn} & $153$ & $0.971$ & $0.654$ & $\phantom{-}0.000$ \\
\lang{jpn} & $153$ & $0.957$ & $0.660$ & $-0.014$ \\
\lang{ita} & $153$ & $0.942$ & $0.654$ & $-0.029$ \\
\lang{fra} & $153$ & $0.928$ & $0.647$ & $-0.043$ \\
\lang{spa} & $153$ & $0.914$ & $0.641$ & $-0.057$ \\
\lang{por} & $153$ & $0.914$ & $0.641$ & $-0.057$ \\
\lang{hin} & $153$ & $0.901$ & $0.621$ & $-0.070$ \\
\lang{deu} & $153$ & $0.887$ & $0.627$ & $-0.084$ \\
\lang{ind} & $153$ & $0.887$ & $0.627$ & $-0.084$ \\
\lang{tur} & $153$ & $0.886$ & $0.641$ & $-0.085$ \\
\bottomrule
\end{tabular}
\caption{J3 agreement with human-majority labels on the LLM-touched slice
of the calibration set, for all eleven in-scope languages. The human pass
rate on this slice is $0.654$ in every language: the English human-majority
label is translation-invariant, so $\kappa$ isolates translation-induced
verdict drift. Every $\Delta\kappa$ sits inside the pre-registered
$\Delta\kappa \leq 0.10$ hazard threshold; the largest drop is \lang{tur}
at $-0.085$, and the non-Latin scripts (\lang{cmn}, \lang{jpn}, \lang{hin})
rank among the strongest, so script family does not predict translation
drift.}
\label{tab:calib_multilingual_llm}
\end{table}

Per-capability multilingual $\kappa$ is reported only for \emph{search}
(Table~\ref{tab:calib_search_multilingual}). We deliberately omit
per-capability $\kappa$ for adaptability, ambiguity and execution under
translation: on those subsets the verdict is deterministic (see
Table~\ref{tab:failure_taxonomy}) and $\kappa$ is language-invariant
by construction, which would document a corpus property rather than
judge quality.

\begin{table}[h]
\centering\small\setlength{\tabcolsep}{4pt}
\begin{tabular}{lccc}
\toprule
\textbf{Language} & Agr. & $\kappa$ & $\Delta$pass \\
\midrule
\lang{eng} & $1.000$ & $1.000$ & $\phantom{-}0.000$ \\
\lang{cmn} & $1.000$ & $1.000$ & $\phantom{-}0.000$ \\
\lang{jpn} & $1.000$ & $1.000$ & $\phantom{-}0.000$ \\
\lang{ita} & $0.979$ & $0.958$ & $\phantom{-}0.000$ \\
\lang{fra} & $0.990$ & $0.979$ & $-0.010$ \\
\lang{spa} & $0.979$ & $0.958$ & $-0.021$ \\
\lang{por} & $0.990$ & $0.979$ & $-0.010$ \\
\lang{hin} & $0.990$ & $0.979$ & $-0.010$ \\
\lang{deu} & $0.979$ & $0.958$ & $-0.021$ \\
\lang{ind} & $0.979$ & $0.958$ & $\phantom{-}0.000$ \\
\lang{tur} & $0.979$ & $0.958$ & $\phantom{-}0.000$ \\
\bottomrule
\end{tabular}
\caption{J3 per-language agreement on the \emph{search} capability
($n=97$), the one capability with a substantial LLM-touched share
($43/97$ in English). Agreement holds at $\kappa \geq 0.958$ across all
eleven languages; \lang{eng}, \lang{jpn} and \lang{cmn} are invariant, and
the largest loss is two verdicts (\lang{spa}, \lang{deu}).}
\label{tab:calib_search_multilingual}
\end{table}

As a same-family bias check on the operational judge, we additionally re-scored a held-out set of \gptoss{} rollouts---used solely for this judge self-preference check, as \gptoss{} is not part of the evaluated agent leaderboard---with both J3 (self) and the cross-family reference J1: the self$-$cross overall \passk{1} difference is negative in all eight complete languages ($\Delta \in [-0.025, -0.014]$), so \gptoss{} shows no self-preference when adjudicating its own family's rollouts---if anything it is marginally stricter than the reference.

\section{Per-language results}
\label{app:per_language}

This appendix expands on the results reported in the main table (Table~\ref{tab:headline}) and the heatmap (Figure~\ref{fig:heatmap_breakdown}) into the full per-agent, per-language, per-capability numbers. 
For each (agent, language, capability) cell we report \passk{1} and \passk{3} with Wilson $95\%$ confidence intervals~\citep{wilson1927}, the all-three-consistency metric \passallk{3}.  Table~\ref{tab:breakdown} reports all four capabilities together, with the four capabilities laid out across the columns.

\onecolumn
\begingroup
\footnotesize\setlength{\tabcolsep}{3pt}\renewcommand{\arraystretch}{0.95}
\begin{longtable}{llcccccccccccc}
\caption{Full per-agent, per-language breakdown across the four reference-oracle capabilities. Each capability reports \passk{1}\,/\,\passk{3}\,/\,\passallk{3} (\%); $n{=}160$ scenarios per cell. \passk{1} is the mean over \emph{valid} runs (agent engaged, \texttt{num\_agent\_events}$>$3; infra non-attempts excluded), \passk{3} is any-of-three over valid runs, and `all' is \passallk{3} (all three valid runs pass). Sub/superscripts on \passk{1}/\passk{3} are Wilson $95\%$ CI bounds.
}\label{tab:breakdown}\\
\toprule
& & \multicolumn{3}{c}{Exec.} & \multicolumn{3}{c}{Search} & \multicolumn{3}{c}{Adapt.} & \multicolumn{3}{c}{Ambig.} \\
\cmidrule(lr){3-5} \cmidrule(lr){6-8} \cmidrule(lr){9-11} \cmidrule(lr){12-14}
Agent & Lang & @1 & @3 & all & @1 & @3 & all & @1 & @3 & all & @1 & @3 & all \\ \midrule
\endfirsthead
\multicolumn{14}{l}{\emph{(Table~\ref{tab:breakdown} continued)}}\\ \toprule
& & \multicolumn{3}{c}{Exec.} & \multicolumn{3}{c}{Search} & \multicolumn{3}{c}{Adapt.} & \multicolumn{3}{c}{Ambig.} \\
\cmidrule(lr){3-5} \cmidrule(lr){6-8} \cmidrule(lr){9-11} \cmidrule(lr){12-14}
Agent & Lang & @1 & @3 & all & @1 & @3 & all & @1 & @3 & all & @1 & @3 & all \\ \midrule
\endhead
\bottomrule \endfoot
\claude & \lang{eng} & 79.4$_{76}^{83}$ & 88.1$_{82}^{92}$ & 69.4 & 84.0$_{80}^{87}$ & 95.6$_{91}^{98}$ & 67.7 & 63.9$_{59}^{68}$ & 75.6$_{68}^{82}$ & 49.4 & 58.0$_{53}^{62}$ & 71.6$_{64}^{78}$ & 43.2 \\
 & \lang{spa} & 70.4$_{66}^{74}$ & 84.4$_{78}^{89}$ & 54.4 & 74.6$_{70}^{78}$ & 89.3$_{84}^{93}$ & 57.9 & 61.8$_{57}^{66}$ & 75.6$_{68}^{82}$ & 43.8 & 49.3$_{45}^{54}$ & 68.8$_{61}^{76}$ & 29.9 \\
 & \lang{deu} & 65.8$_{61}^{70}$ & 79.4$_{72}^{85}$ & 48.1 & 58.4$_{54}^{63}$ & 78.0$_{71}^{84}$ & 34.6 & 58.0$_{53}^{63}$ & 70.0$_{62}^{77}$ & 47.5 & 42.2$_{38}^{47}$ & 58.3$_{50}^{66}$ & 28.2 \\
 & \lang{fra} & 66.7$_{62}^{71}$ & 83.1$_{77}^{88}$ & 47.5 & 65.1$_{60}^{70}$ & 81.1$_{74}^{86}$ & 50.3 & 60.5$_{56}^{65}$ & 72.5$_{65}^{79}$ & 46.2 & 42.0$_{38}^{47}$ & 61.1$_{53}^{68}$ & 24.8 \\
 & \lang{ita} & 59.8$_{55}^{64}$ & 76.2$_{69}^{82}$ & 41.2 & 69.0$_{64}^{73}$ & 81.0$_{74}^{86}$ & 55.7 & 62.2$_{58}^{66}$ & 74.4$_{67}^{81}$ & 48.8 & 35.3$_{31}^{40}$ & 54.1$_{46}^{62}$ & 18.5 \\
 & \lang{por} & 64.9$_{60}^{70}$ & 81.9$_{75}^{87}$ & 46.2 & 70.6$_{66}^{75}$ & 84.9$_{79}^{90}$ & 52.2 & 54.9$_{50}^{59}$ & 73.8$_{66}^{80}$ & 35.0 & 41.3$_{37}^{46}$ & 60.3$_{52}^{68}$ & 23.7 \\
 & \lang{ind} & 69.5$_{65}^{73}$ & 83.1$_{77}^{88}$ & 55.6 & 69.4$_{65}^{73}$ & 86.2$_{80}^{91}$ & 50.3 & 61.6$_{57}^{66}$ & 73.8$_{66}^{80}$ & 45.6 & 44.0$_{40}^{49}$ & 60.9$_{53}^{68}$ & 26.9 \\
 & \lang{tur} & 61.9$_{57}^{67}$ & 77.5$_{70}^{83}$ & 45.6 & 68.3$_{64}^{72}$ & 83.0$_{76}^{88}$ & 52.2 & 52.0$_{48}^{56}$ & 72.5$_{65}^{79}$ & 31.9 & 42.4$_{38}^{47}$ & 66.7$_{59}^{74}$ & 19.2 \\
 & \lang{cmn} & 62.4$_{58}^{67}$ & 75.0$_{68}^{81}$ & 49.4 & 66.0$_{62}^{70}$ & 81.8$_{75}^{87}$ & 47.2 & 56.7$_{52}^{61}$ & 73.1$_{66}^{79}$ & 38.1 & 40.5$_{36}^{45}$ & 55.8$_{48}^{63}$ & 28.8 \\
 & \lang{jpn} & 56.1$_{52}^{61}$ & 70.0$_{62}^{77}$ & 41.2 & 65.3$_{61}^{69}$ & 83.0$_{76}^{88}$ & 47.8 & 55.3$_{51}^{60}$ & 73.0$_{66}^{79}$ & 33.3 & 35.9$_{32}^{40}$ & 51.9$_{44}^{60}$ & 19.0 \\
 & \lang{hin} & 53.7$_{49}^{58}$ & 77.5$_{70}^{83}$ & 30.0 & 63.8$_{59}^{68}$ & 81.8$_{75}^{87}$ & 44.0 & 54.8$_{50}^{60}$ & 71.9$_{64}^{78}$ & 36.2 & 40.5$_{36}^{45}$ & 59.2$_{51}^{67}$ & 21.0 \\
\midrule
\gptfive & \lang{eng} & 63.0$_{59}^{67}$ & 81.2$_{74}^{87}$ & 44.4 & 82.1$_{78}^{85}$ & 92.4$_{87}^{96}$ & 70.1 & 39.2$_{35}^{44}$ & 58.8$_{51}^{66}$ & 21.9 & 25.6$_{22}^{30}$ & 41.4$_{34}^{49}$ & 12.7 \\
 & \lang{spa} & 51.9$_{47}^{56}$ & 69.4$_{62}^{76}$ & 33.1 & 74.8$_{71}^{79}$ & 87.3$_{81}^{92}$ & 61.8 & 34.8$_{31}^{39}$ & 51.2$_{44}^{59}$ & 18.1 & 19.0$_{16}^{23}$ & 33.3$_{26}^{41}$ & 8.2 \\
 & \lang{deu} & 52.5$_{48}^{57}$ & 68.8$_{61}^{75}$ & 32.5 & 72.1$_{68}^{76}$ & 85.4$_{79}^{90}$ & 56.1 & 36.0$_{32}^{40}$ & 53.8$_{46}^{61}$ & 19.4 & 21.4$_{18}^{25}$ & 35.4$_{28}^{43}$ & 8.9 \\
 & \lang{fra} & 49.9$_{45}^{54}$ & 65.0$_{57}^{72}$ & 31.9 & 75.1$_{71}^{79}$ & 88.6$_{83}^{93}$ & 61.4 & 35.6$_{31}^{40}$ & 52.5$_{45}^{60}$ & 17.5 & 21.1$_{18}^{25}$ & 34.4$_{27}^{42}$ & 9.6 \\
 & \lang{ita} & 49.8$_{45}^{54}$ & 70.0$_{62}^{77}$ & 28.7 & 73.6$_{69}^{77}$ & 86.7$_{81}^{91}$ & 57.6 & 38.5$_{34}^{43}$ & 57.5$_{50}^{65}$ & 18.8 & 17.3$_{14}^{21}$ & 29.1$_{23}^{37}$ & 7.6 \\
 & \lang{por} & 53.7$_{49}^{58}$ & 73.8$_{66}^{80}$ & 34.4 & 71.4$_{67}^{75}$ & 85.3$_{79}^{90}$ & 56.4 & 34.5$_{30}^{39}$ & 51.2$_{44}^{59}$ & 17.5 & 19.6$_{16}^{23}$ & 33.1$_{26}^{41}$ & 8.9 \\
 & \lang{ind} & 52.8$_{48}^{57}$ & 70.6$_{63}^{77}$ & 30.6 & 71.9$_{68}^{76}$ & 84.6$_{78}^{89}$ & 55.8 & 35.0$_{31}^{39}$ & 54.4$_{47}^{62}$ & 16.9 & 19.7$_{16}^{24}$ & 32.5$_{26}^{40}$ & 9.6 \\
 & \lang{tur} & 53.1$_{49}^{58}$ & 73.8$_{66}^{80}$ & 35.0 & 66.8$_{62}^{71}$ & 78.5$_{71}^{84}$ & 52.5 & 35.6$_{31}^{40}$ & 52.5$_{45}^{60}$ & 19.4 & 20.9$_{17}^{25}$ & 34.6$_{28}^{42}$ & 10.7 \\
 & \lang{cmn} & 41.8$_{37}^{46}$ & 57.2$_{49}^{65}$ & 24.5 & 65.5$_{61}^{70}$ & 80.3$_{73}^{86}$ & 48.4 & 33.3$_{29}^{38}$ & 48.8$_{41}^{56}$ & 17.5 & 18.2$_{15}^{22}$ & 29.9$_{23}^{38}$ & 10.2 \\
 & \lang{jpn} & 40.5$_{36}^{45}$ & 58.1$_{50}^{65}$ & 24.4 & 61.1$_{57}^{65}$ & 77.6$_{70}^{83}$ & 42.3 & 24.9$_{21}^{29}$ & 43.1$_{36}^{51}$ & 10.0 & 15.6$_{13}^{19}$ & 26.8$_{20}^{34}$ & 8.3 \\
 & \lang{hin} & 45.4$_{41}^{50}$ & 64.2$_{56}^{71}$ & 24.5 & 66.0$_{62}^{70}$ & 80.8$_{74}^{86}$ & 53.2 & 31.2$_{27}^{36}$ & 49.4$_{42}^{57}$ & 14.4 & 16.4$_{13}^{20}$ & 29.1$_{23}^{37}$ & 5.7 \\
\midrule
\gemini & \lang{eng} & 48.4$_{44}^{53}$ & 66.2$_{59}^{73}$ & 28.8 & 67.6$_{63}^{72}$ & 86.6$_{80}^{91}$ & 47.1 & 31.2$_{27}^{36}$ & 51.2$_{44}^{59}$ & 12.5 & 26.4$_{23}^{31}$ & 43.7$_{36}^{51}$ & 12.0 \\
 & \lang{spa} & 37.9$_{34}^{42}$ & 56.2$_{49}^{64}$ & 20.0 & 62.4$_{58}^{67}$ & 80.5$_{74}^{86}$ & 43.4 & 29.6$_{26}^{34}$ & 46.2$_{39}^{54}$ & 13.1 & 21.2$_{18}^{25}$ & 34.8$_{28}^{43}$ & 10.1 \\
 & \lang{deu} & 38.4$_{34}^{43}$ & 59.4$_{52}^{67}$ & 19.4 & 58.8$_{54}^{63}$ & 79.7$_{73}^{85}$ & 38.6 & 31.2$_{27}^{36}$ & 50.6$_{43}^{58}$ & 12.5 & 20.8$_{17}^{25}$ & 37.3$_{30}^{45}$ & 6.3 \\
 & \lang{fra} & 40.1$_{36}^{45}$ & 55.6$_{48}^{63}$ & 25.6 & 60.0$_{56}^{64}$ & 77.2$_{70}^{83}$ & 43.0 & 31.0$_{27}^{35}$ & 51.2$_{44}^{59}$ & 11.9 & 18.8$_{15}^{23}$ & 30.2$_{24}^{38}$ & 8.2 \\
 & \lang{ita} & 36.0$_{32}^{40}$ & 53.8$_{46}^{61}$ & 18.1 & 60.1$_{56}^{64}$ & 79.1$_{72}^{85}$ & 37.3 & 29.8$_{26}^{34}$ & 51.2$_{44}^{59}$ & 10.0 & 18.9$_{16}^{23}$ & 32.1$_{25}^{40}$ & 8.2 \\
 & \lang{por} & 39.2$_{35}^{44}$ & 60.0$_{52}^{67}$ & 20.6 & 59.7$_{55}^{64}$ & 77.4$_{70}^{83}$ & 40.9 & 30.8$_{27}^{35}$ & 53.1$_{45}^{61}$ & 11.2 & 18.7$_{15}^{22}$ & 31.6$_{25}^{39}$ & 8.9 \\
 & \lang{ind} & 38.2$_{34}^{43}$ & 58.8$_{51}^{66}$ & 18.8 & 62.2$_{58}^{66}$ & 80.5$_{74}^{86}$ & 42.1 & 31.5$_{27}^{36}$ & 48.8$_{41}^{56}$ & 16.9 & 16.5$_{13}^{20}$ & 28.9$_{22}^{36}$ & 7.5 \\
 & \lang{tur} & 37.7$_{33}^{42}$ & 54.4$_{47}^{62}$ & 21.9 & 53.2$_{49}^{58}$ & 75.8$_{69}^{82}$ & 31.2 & 25.5$_{22}^{30}$ & 46.2$_{39}^{54}$ & 8.8 & 17.0$_{14}^{21}$ & 28.5$_{22}^{36}$ & 9.5 \\
 & \lang{cmn} & 41.8$_{37}^{46}$ & 60.6$_{53}^{68}$ & 23.1 & 56.9$_{52}^{61}$ & 77.2$_{70}^{83}$ & 34.8 & 28.6$_{25}^{33}$ & 50.6$_{43}^{58}$ & 7.5 & 14.9$_{12}^{18}$ & 24.1$_{18}^{31}$ & 8.2 \\
 & \lang{jpn} & 36.3$_{32}^{41}$ & 53.1$_{45}^{61}$ & 19.4 & 54.9$_{50}^{59}$ & 74.8$_{68}^{81}$ & 34.6 & 24.8$_{21}^{29}$ & 43.1$_{36}^{51}$ & 8.1 & 16.5$_{13}^{20}$ & 26.6$_{20}^{34}$ & 7.0 \\
 & \lang{hin} & 31.9$_{28}^{36}$ & 49.4$_{42}^{57}$ & 16.9 & 58.4$_{54}^{63}$ & 79.0$_{72}^{85}$ & 36.9 & 25.0$_{21}^{29}$ & 47.5$_{40}^{55}$ & 8.1 & 14.3$_{11}^{18}$ & 24.5$_{18}^{32}$ & 6.3 \\
\midrule
\kimi & \lang{eng} & 51.7$_{47}^{56}$ & 73.8$_{66}^{80}$ & 28.7 & 81.5$_{78}^{85}$ & 94.3$_{90}^{97}$ & 63.9 & 35.7$_{32}^{40}$ & 55.0$_{47}^{63}$ & 18.8 & 25.5$_{22}^{30}$ & 42.8$_{35}^{51}$ & 9.4 \\
 & \lang{spa} & 36.9$_{33}^{41}$ & 57.5$_{50}^{65}$ & 18.1 & 73.4$_{69}^{77}$ & 91.2$_{86}^{95}$ & 53.5 & 31.5$_{28}^{36}$ & 48.1$_{41}^{56}$ & 17.5 & 20.5$_{17}^{24}$ & 36.7$_{30}^{44}$ & 8.2 \\
 & \lang{deu} & 37.1$_{33}^{42}$ & 55.6$_{48}^{63}$ & 19.4 & 64.9$_{60}^{69}$ & 82.3$_{76}^{87}$ & 43.0 & 31.4$_{27}^{36}$ & 48.1$_{41}^{56}$ & 15.6 & 17.3$_{14}^{21}$ & 28.5$_{22}^{36}$ & 7.6 \\
 & \lang{fra} & 36.9$_{33}^{41}$ & 56.2$_{49}^{64}$ & 18.8 & 70.9$_{67}^{75}$ & 85.5$_{79}^{90}$ & 51.6 & 31.5$_{28}^{36}$ & 46.9$_{39}^{55}$ & 17.5 & 21.3$_{18}^{25}$ & 35.0$_{28}^{43}$ & 7.6 \\
 & \lang{ita} & 38.3$_{34}^{43}$ & 58.1$_{50}^{65}$ & 18.1 & 69.9$_{66}^{74}$ & 87.3$_{81}^{92}$ & 48.7 & 31.9$_{28}^{36}$ & 48.8$_{41}^{56}$ & 15.0 & 19.5$_{16}^{23}$ & 30.6$_{24}^{38}$ & 10.0 \\
 & \lang{por} & 38.5$_{34}^{43}$ & 59.4$_{52}^{67}$ & 20.6 & 72.3$_{68}^{76}$ & 88.1$_{82}^{92}$ & 53.5 & 26.1$_{22}^{30}$ & 42.5$_{35}^{50}$ & 13.8 & 16.8$_{14}^{20}$ & 28.7$_{22}^{36}$ & 6.9 \\
 & \lang{ind} & 33.8$_{30}^{38}$ & 50.6$_{43}^{58}$ & 16.2 & 70.7$_{66}^{75}$ & 89.9$_{84}^{94}$ & 46.2 & 25.5$_{22}^{30}$ & 41.2$_{34}^{49}$ & 10.6 & 16.1$_{13}^{20}$ & 29.4$_{23}^{37}$ & 5.0 \\
 & \lang{tur} & 36.7$_{32}^{41}$ & 62.5$_{55}^{70}$ & 13.8 & 63.1$_{59}^{67}$ & 79.0$_{72}^{85}$ & 44.6 & 26.7$_{23}^{31}$ & 45.6$_{38}^{53}$ & 9.4 & 17.2$_{14}^{21}$ & 30.8$_{24}^{38}$ & 5.0 \\
 & \lang{cmn} & 35.0$_{31}^{39}$ & 53.1$_{45}^{61}$ & 15.6 & 64.6$_{60}^{69}$ & 83.0$_{76}^{88}$ & 45.3 & 26.3$_{23}^{30}$ & 40.0$_{33}^{48}$ & 13.8 & 21.0$_{18}^{25}$ & 34.0$_{27}^{42}$ & 8.8 \\
 & \lang{jpn} & 35.4$_{31}^{40}$ & 53.8$_{46}^{61}$ & 16.2 & 63.4$_{59}^{68}$ & 79.6$_{73}^{85}$ & 43.9 & 25.9$_{22}^{30}$ & 43.1$_{36}^{51}$ & 10.6 & 16.7$_{14}^{20}$ & 30.4$_{24}^{38}$ & 5.7 \\
 & \lang{hin} & 33.8$_{30}^{38}$ & 58.1$_{50}^{65}$ & 15.0 & 61.9$_{57}^{66}$ & 82.4$_{76}^{88}$ & 39.0 & 24.0$_{20}^{28}$ & 40.3$_{33}^{48}$ & 10.1 & 14.1$_{11}^{18}$ & 26.9$_{21}^{34}$ & 5.6 \\
\midrule
\gemma & \lang{eng} & 50.3$_{46}^{55}$ & 67.5$_{60}^{74}$ & 34.4 & 70.9$_{67}^{75}$ & 85.6$_{79}^{90}$ & 50.0 & 45.2$_{41}^{50}$ & 63.1$_{55}^{70}$ & 26.2 & 23.8$_{20}^{28}$ & 36.9$_{30}^{45}$ & 10.6 \\
 & \lang{spa} & 41.9$_{38}^{46}$ & 56.9$_{49}^{64}$ & 25.6 & 63.6$_{59}^{68}$ & 78.1$_{71}^{84}$ & 41.2 & 39.0$_{35}^{43}$ & 57.5$_{50}^{65}$ & 23.1 & 18.3$_{15}^{22}$ & 28.1$_{22}^{36}$ & 8.8 \\
 & \lang{deu} & 45.6$_{41}^{50}$ & 58.8$_{51}^{66}$ & 32.5 & 59.1$_{55}^{64}$ & 76.2$_{69}^{82}$ & 40.0 & 37.3$_{33}^{42}$ & 57.5$_{50}^{65}$ & 19.4 & 18.5$_{15}^{22}$ & 31.9$_{25}^{39}$ & 6.9 \\
 & \lang{fra} & 42.1$_{38}^{47}$ & 59.4$_{52}^{67}$ & 23.8 & 62.2$_{58}^{67}$ & 75.6$_{68}^{82}$ & 48.1 & 41.2$_{37}^{46}$ & 56.2$_{49}^{64}$ & 27.5 & 18.9$_{16}^{23}$ & 31.2$_{25}^{39}$ & 8.8 \\
 & \lang{ita} & 43.4$_{39}^{48}$ & 60.0$_{52}^{67}$ & 28.8 & 63.6$_{59}^{68}$ & 76.2$_{69}^{82}$ & 42.5 & 41.8$_{38}^{46}$ & 58.1$_{50}^{65}$ & 26.9 & 16.8$_{14}^{20}$ & 28.1$_{22}^{36}$ & 8.1 \\
 & \lang{por} & 47.3$_{43}^{52}$ & 62.5$_{55}^{70}$ & 31.2 & 60.8$_{56}^{65}$ & 71.2$_{64}^{78}$ & 40.6 & 35.5$_{31}^{40}$ & 50.6$_{43}^{58}$ & 20.0 & 18.6$_{15}^{22}$ & 28.1$_{22}^{36}$ & 8.8 \\
 & \lang{ind} & 45.2$_{41}^{50}$ & 61.2$_{54}^{68}$ & 27.5 & 60.2$_{56}^{65}$ & 75.0$_{68}^{81}$ & 37.5 & 40.4$_{36}^{45}$ & 59.4$_{52}^{67}$ & 20.6 & 18.5$_{15}^{22}$ & 31.2$_{25}^{39}$ & 8.1 \\
 & \lang{tur} & 40.5$_{36}^{45}$ & 56.9$_{49}^{64}$ & 25.0 & 52.5$_{48}^{57}$ & 65.0$_{57}^{72}$ & 36.9 & 35.8$_{32}^{40}$ & 50.6$_{43}^{58}$ & 20.6 & 18.3$_{15}^{22}$ & 30.6$_{24}^{38}$ & 6.9 \\
 & \lang{cmn} & 35.2$_{31}^{40}$ & 49.4$_{42}^{57}$ & 20.0 & 46.3$_{42}^{51}$ & 63.1$_{55}^{70}$ & 26.2 & 36.2$_{32}^{41}$ & 48.1$_{41}^{56}$ & 23.1 & 18.4$_{15}^{22}$ & 25.6$_{19}^{33}$ & 10.0 \\
 & \lang{jpn} & 31.6$_{28}^{36}$ & 41.9$_{35}^{50}$ & 21.2 & 42.3$_{38}^{47}$ & 59.4$_{52}^{67}$ & 21.2 & 30.5$_{27}^{35}$ & 47.5$_{40}^{55}$ & 11.9 & 13.1$_{10}^{16}$ & 20.0$_{15}^{27}$ & 7.5 \\
 & \lang{hin} & 28.5$_{25}^{33}$ & 45.0$_{37}^{53}$ & 14.4 & 39.9$_{35}^{44}$ & 55.0$_{47}^{63}$ & 20.6 & 33.7$_{30}^{38}$ & 48.8$_{41}^{56}$ & 18.1 & 13.0$_{10}^{16}$ & 20.0$_{15}^{27}$ & 5.6 \\
\midrule
\qwenthreesixathreeb & \lang{eng} & 49.0$_{45}^{53}$ & 68.8$_{61}^{75}$ & 28.1 & 63.2$_{59}^{67}$ & 86.2$_{80}^{91}$ & 34.4 & 32.1$_{28}^{36}$ & 48.1$_{41}^{56}$ & 16.2 & 11.7$_{9}^{15}$ & 23.1$_{17}^{30}$ & 3.1 \\
 & \lang{spa} & 26.5$_{23}^{31}$ & 46.2$_{39}^{54}$ & 10.6 & 41.4$_{37}^{46}$ & 69.4$_{62}^{76}$ & 16.9 & 22.3$_{19}^{26}$ & 37.5$_{30}^{45}$ & 5.6 & 6.4$_{5}^{9}$ & 14.4$_{10}^{21}$ & 1.2 \\
 & \lang{deu} & 28.0$_{24}^{32}$ & 48.1$_{41}^{56}$ & 11.2 & 36.8$_{33}^{41}$ & 65.6$_{58}^{73}$ & 10.0 & 17.6$_{14}^{21}$ & 36.9$_{30}^{45}$ & 2.5 & 6.8$_{5}^{9}$ & 11.9$_{8}^{18}$ & 3.1 \\
 & \lang{fra} & 27.8$_{24}^{32}$ & 44.4$_{37}^{52}$ & 10.6 & 44.7$_{40}^{49}$ & 71.2$_{64}^{78}$ & 16.9 & 20.3$_{17}^{24}$ & 35.0$_{28}^{43}$ & 6.2 & 8.4$_{6}^{11}$ & 15.0$_{10}^{21}$ & 2.5 \\
 & \lang{ita} & 28.5$_{25}^{33}$ & 48.1$_{41}^{56}$ & 11.9 & 40.9$_{36}^{45}$ & 66.9$_{59}^{74}$ & 13.1 & 19.4$_{16}^{23}$ & 33.1$_{26}^{41}$ & 6.9 & 8.2$_{6}^{11}$ & 13.8$_{9}^{20}$ & 4.4 \\
 & \lang{por} & 28.5$_{25}^{33}$ & 47.5$_{40}^{55}$ & 10.0 & 40.4$_{36}^{45}$ & 68.8$_{61}^{75}$ & 11.9 & 19.9$_{17}^{24}$ & 32.5$_{26}^{40}$ & 8.8 & 8.5$_{6}^{11}$ & 15.6$_{11}^{22}$ & 1.9 \\
 & \lang{ind} & 25.7$_{22}^{30}$ & 42.5$_{35}^{50}$ & 9.4 & 40.2$_{36}^{45}$ & 66.9$_{59}^{74}$ & 13.8 & 21.3$_{18}^{25}$ & 40.6$_{33}^{48}$ & 4.4 & 7.2$_{5}^{10}$ & 13.1$_{9}^{19}$ & 1.9 \\
 & \lang{tur} & 22.7$_{19}^{27}$ & 40.6$_{33}^{48}$ & 8.1 & 35.7$_{31}^{40}$ & 63.1$_{55}^{70}$ & 11.9 & 15.8$_{13}^{19}$ & 29.4$_{23}^{37}$ & 3.8 & 7.2$_{5}^{10}$ & 14.4$_{10}^{21}$ & 1.2 \\
 & \lang{cmn} & 27.4$_{24}^{32}$ & 41.9$_{35}^{50}$ & 11.9 & 38.7$_{34}^{43}$ & 63.1$_{55}^{70}$ & 13.1 & 17.5$_{14}^{21}$ & 34.4$_{27}^{42}$ & 3.1 & 8.8$_{7}^{12}$ & 13.1$_{9}^{19}$ & 4.4 \\
 & \lang{jpn} & 17.8$_{15}^{22}$ & 30.0$_{23}^{38}$ & 7.5 & 35.7$_{31}^{40}$ & 61.9$_{54}^{69}$ & 6.9 & 12.3$_{10}^{16}$ & 23.1$_{17}^{30}$ & 5.0 & 6.7$_{5}^{9}$ & 13.8$_{9}^{20}$ & 2.5 \\
 & \lang{hin} & 15.9$_{13}^{19}$ & 28.1$_{22}^{36}$ & 5.0 & 31.6$_{27}^{36}$ & 58.8$_{51}^{66}$ & 8.1 & 8.1$_{6}^{11}$ & 15.6$_{11}^{22}$ & 1.9 & 5.3$_{4}^{8}$ & 10.6$_{7}^{16}$ & 0.0 \\
\midrule
\qwenthreesixdense & \lang{eng} & 54.8$_{50}^{59}$ & 77.5$_{70}^{83}$ & 26.9 & 58.4$_{54}^{63}$ & 79.4$_{72}^{85}$ & 31.2 & 39.6$_{35}^{44}$ & 61.2$_{54}^{68}$ & 15.0 & 20.0$_{17}^{24}$ & 30.6$_{24}^{38}$ & 6.2 \\
 & \lang{spa} & 38.3$_{34}^{43}$ & 59.4$_{52}^{67}$ & 15.0 & 49.9$_{45}^{55}$ & 76.2$_{69}^{82}$ & 20.6 & 38.8$_{35}^{43}$ & 60.0$_{52}^{67}$ & 15.6 & 15.5$_{12}^{19}$ & 25.6$_{19}^{33}$ & 4.4 \\
 & \lang{deu} & 39.0$_{35}^{44}$ & 59.4$_{52}^{67}$ & 18.1 & 46.3$_{42}^{51}$ & 66.9$_{59}^{74}$ & 20.0 & 36.5$_{32}^{41}$ & 54.4$_{47}^{62}$ & 14.4 & 14.3$_{11}^{18}$ & 24.4$_{18}^{32}$ & 5.0 \\
 & \lang{fra} & 40.3$_{36}^{45}$ & 58.8$_{51}^{66}$ & 16.2 & 47.5$_{43}^{52}$ & 73.8$_{66}^{80}$ & 18.8 & 39.3$_{35}^{44}$ & 59.4$_{52}^{67}$ & 16.2 & 13.1$_{10}^{17}$ & 22.5$_{17}^{30}$ & 5.6 \\
 & \lang{ita} & 40.9$_{36}^{45}$ & 60.0$_{52}^{67}$ & 18.1 & 50.9$_{46}^{56}$ & 74.4$_{67}^{81}$ & 22.5 & 36.9$_{33}^{41}$ & 56.2$_{49}^{64}$ & 17.5 & 16.0$_{13}^{20}$ & 29.4$_{23}^{37}$ & 5.6 \\
 & \lang{por} & 39.5$_{35}^{44}$ & 58.8$_{51}^{66}$ & 18.1 & 50.5$_{46}^{55}$ & 69.4$_{62}^{76}$ & 21.9 & 32.4$_{28}^{37}$ & 50.6$_{43}^{58}$ & 10.6 & 15.5$_{12}^{19}$ & 27.5$_{21}^{35}$ & 5.0 \\
 & \lang{ind} & 39.4$_{35}^{44}$ & 61.9$_{54}^{69}$ & 14.4 & 48.3$_{44}^{53}$ & 70.6$_{63}^{77}$ & 18.1 & 32.3$_{28}^{37}$ & 51.9$_{44}^{59}$ & 11.9 & 12.4$_{10}^{16}$ & 21.9$_{16}^{29}$ & 3.1 \\
 & \lang{tur} & 33.3$_{29}^{38}$ & 54.4$_{47}^{62}$ & 13.1 & 44.6$_{40}^{49}$ & 64.4$_{57}^{71}$ & 18.8 & 32.6$_{29}^{37}$ & 53.1$_{45}^{61}$ & 13.1 & 15.9$_{13}^{20}$ & 28.8$_{22}^{36}$ & 5.0 \\
 & \lang{cmn} & 33.3$_{29}^{38}$ & 53.8$_{46}^{61}$ & 15.0 & 43.5$_{39}^{48}$ & 68.8$_{61}^{75}$ & 13.1 & 33.5$_{29}^{38}$ & 53.1$_{45}^{61}$ & 11.2 & 15.8$_{13}^{19}$ & 27.5$_{21}^{35}$ & 3.1 \\
 & \lang{jpn} & 27.5$_{24}^{32}$ & 43.1$_{36}^{51}$ & 11.2 & 40.6$_{36}^{45}$ & 66.2$_{59}^{73}$ & 11.2 & 23.4$_{20}^{28}$ & 38.8$_{32}^{46}$ & 8.1 & 11.6$_{9}^{15}$ & 19.4$_{14}^{26}$ & 4.4 \\
 & \lang{hin} & 22.8$_{19}^{27}$ & 42.5$_{35}^{50}$ & 3.1 & 37.2$_{33}^{42}$ & 63.8$_{56}^{71}$ & 10.0 & 24.3$_{21}^{28}$ & 41.9$_{35}^{50}$ & 7.5 & 9.5$_{7}^{13}$ & 16.9$_{12}^{23}$ & 3.8 \\
\end{longtable}
\endgroup

\section{\qwenthreefive{} scale ladder: per-language cells}
\label{app:qwen_ladder}

Table~\ref{tab:qwen_ladder_lang} expands the pooled scale ladder of
\S\ref{sec:analysis-scale} (Table~\ref{tab:qwen_ladder}) into per-language
\passk{1} for the four \qwenthreefive{} sizes, averaged over the four
reference-oracle-backed capabilities (\texttt{time} excluded). Each cell pools
$4{\times}160{=}640$ scenarios per (size, language). English is the strongest
column at every size and \lang{hin} the weakest, with \lang{jpn} close behind;
the English-minus-target gap (\S\ref{sec:analysis-scale}) is visible as the
spread between the first column and the rest and \emph{widens} with scale.

\begin{table*}[ht]
  \centering \small
  \setlength{\tabcolsep}{4pt}
  \begin{tabular}{l ccccccccccc}
    \toprule
    Rung & \lang{eng} & \lang{cmn} & \lang{deu} & \lang{fra} & \lang{hin} & \lang{ind} & \lang{ita} & \lang{jpn} & \lang{por} & \lang{spa} & \lang{tur} \\
    \midrule
    \model{35B-A3B}   & 17.8 &  7.7 & 11.6 & 12.5 &  3.0 & 14.7 & 10.2 &  4.7 & 11.2 & 13.9 &  7.8 \\
    \model{122B-A10B} & 19.7 & 11.9 & 13.1 & 14.5 &  6.6 & 14.8 & 14.2 &  7.0 & 14.4 & 15.5 &  9.8 \\
    \model{27B}       & 27.5 & 15.9 & 18.9 & 21.4 &  7.3 & 19.2 & 20.3 & \textbf{12.5} & 19.4 & 22.5 & 13.6 \\
    \model{397B-A17B} & \textbf{31.6} & \textbf{16.2} & \textbf{21.1} & \textbf{22.5} & \textbf{9.2} & \textbf{21.6} & \textbf{23.9} & 10.0 & \textbf{23.0} & \textbf{23.1} & \textbf{17.7} \\
    \bottomrule
  \end{tabular}
  \caption{\qwenthreefive{} scale ladder, per-language \passk{1} (\%) averaged
  over the four capabilities (execution, search, adaptability, ambiguity;
  \texttt{time} excluded), pooling $4{\times}160{=}640$ scenarios per (size,
  language). Best per column in \textbf{bold}: \model{397B-A17B} leads every
  language except \lang{jpn} (won by the dense \model{27B}). English is the
  ceiling and \lang{hin}/\lang{jpn} the floor at every size. Protocol and
  caveats are as in Table~\ref{tab:qwen_ladder}.}
  \label{tab:qwen_ladder_lang}
\end{table*}

\paragraph{Reasoning budget (thinking effort).} Increasing the agent's thinking
effort from low to high does not improve agentic \passk{1} on this benchmark. On
the largest \model{397B-A17B} size, a high-effort rerun matches the low-effort
ladder to within about two points per capability and is, if anything, marginally
\emph{lower} (adaptability $17.0$ vs $17.4$; ambiguity $5.2$ vs $7.5$; execution
tracking similarly). We therefore report the ladder at
\texttt{thinking\_effort}$=$low, matching the headline open-weights
configuration.

\section{Reasoning-effort ladder}
\label{app:effort}

Given budget constraints, we run proprietary models at the provider default reasoning effort.
In the age of test-time scaling~\citep{snell2024scalingllmtesttimecompute}, it's only fair to ask whether the cross-lingual gap we observe (and report in \S\ref{sec:results}) is an artefact of an insufficient \emph{reasoning budget} rather than a genuine capability deficit.  
We test this directly for \claude{}, the strongest frontier agent, sweeping two of the highest \emph{extended thinking}\footnote{See \href{https://platform.claude.com/docs/en/build-with-claude/extended-thinking}{Anthropic's extended-thinking documentation}.} settings (\texttt{high} and \texttt{xhigh}) over three
target languages, \lang{ita}, \lang{jpn} and \lang{tur}, spanning a
Romance-Latin, a Japonic mixed-script and a Turkic agglutinative language and
all four capabilities, and study how \passk{3} varies with reasoning effort.

\begin{figure*}[ht]
\centering
\includegraphics[width=0.75\textwidth]{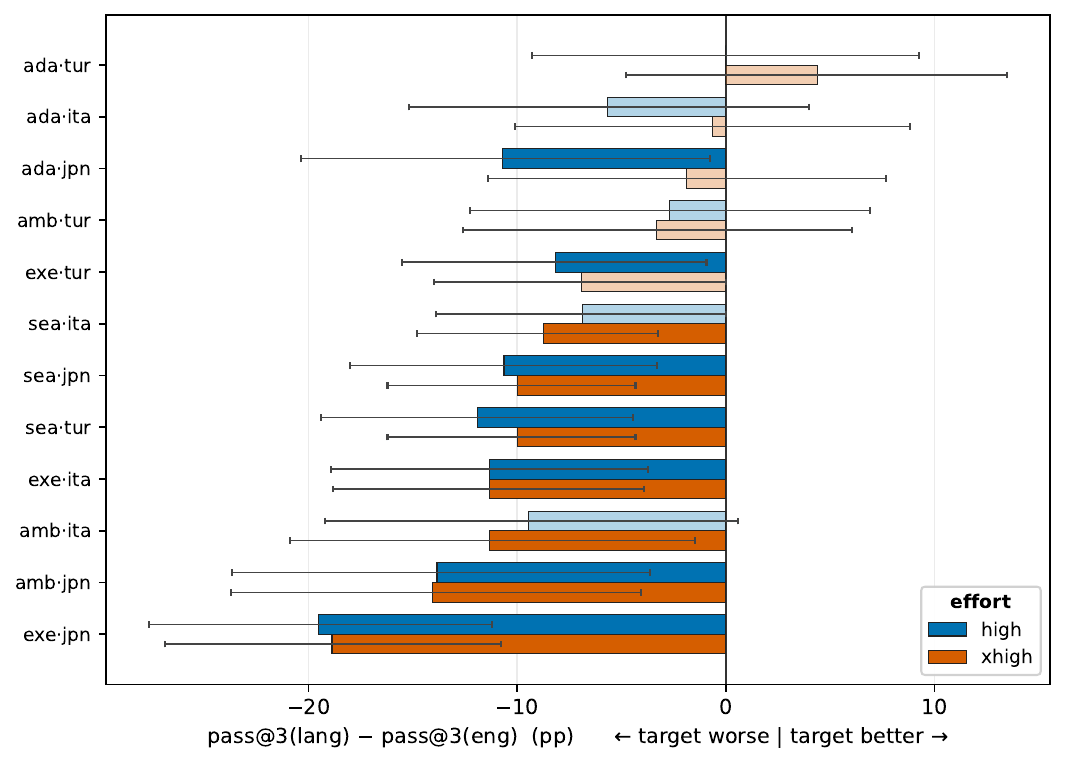}
\caption{Target-minus-English \passk{3} marginal gaps for \claude{} at both tested effort settings (\texttt{high} and \texttt{xhigh}), one grouped pair per (capability, language) cell, sorted by the \texttt{xhigh} gap. Solid bars mark cells whose $95\%$ Wilson--Newcombe CI on the marginal gap excludes zero; washed bars cross zero.}
\label{fig:effort-gap}
\end{figure*}

Note that exposing a controllable effort setting required re-running \claude{} under a separate configuration; we therefore compare only \emph{within} the ladder, reporting target-minus-English marginal gaps at a fixed effort. 
Table~\ref{tab:effort} reports these gaps at both tested settings.
The regression does not close: seven of the twelve target-language $\times$ capability cells retain a significant negative gap, the mean gap over all twelve cells is $-7.8$ \passk{3} points, and the largest single cell---execution on Japanese, $-18.9$ points with a $95\%$ confidence interval of $[-26.9, -10.8]$---is essentially unchanged from its headline value at the very setting the objection predicts should close it (Figure~\ref{fig:effort-gap}). 
Raising the budget from \texttt{high} to \texttt{xhigh} narrows the mean marginal gap only marginally (from $-9.2$ to $-7.8$ points) and leaves the sign pattern unchanged, with seven significant negative cells at either setting.

The residual gap is, however, capability-dependent.
Execution, the largest gap, is the least responsive to effort (mean $-12.4$ points at \texttt{xhigh}), while adaptability, whose headline gap is already small, is neutralised (mean $+0.6$ points, with Turkish even overtaking English by $+4.4$); Japanese is the single cell where raising the budget turns a significant gap non-significant, from $-10.7$ at \texttt{high} to $-1.9$ at \texttt{xhigh}. 
Search and ambiguity sit in between (both mean $-9.6$).
We therefore read reasoning effort as a second-order modulator of the cross-lingual gap rather than its cause: at the effort levels we can measure, the majority of the multilingual regression persists, and it is concentrated on the same execution and ambiguity axes that we identified (\S\ref{sec:analysis-capability}, \S\ref{sec:analysis-checker}).

\begin{table*}[ht]
  \centering \small
  \setlength{\tabcolsep}{4pt}
  \begin{tabular}{l l r r c r r c}
    \toprule
    & & \multicolumn{3}{c}{\texttt{high}} & \multicolumn{3}{c}{\texttt{xhigh}} \\
    \cmidrule(lr){3-5}\cmidrule(lr){6-8}
    Capability & Lang & \passk{3} (\%) & Gap (\%) & 95\% CI & \passk{3} (\%) & Gap (\%) & 95\% CI \\
    \midrule
    \multirow{4}{*}{Execution}
      & \lang{eng} & $91.8$ & -- & & $92.5$ & -- & \\
      & \lang{ita} & $80.5$ & $\mathbf{-11.3}$ & $[-18.9, -3.7]$ & $81.1$ & $\mathbf{-11.3}$ & $[-18.8, -3.9]$ \\
      & \lang{jpn} & $72.3$ & $\mathbf{-19.5}$ & $[-27.6, -11.2]$ & $73.6$ & $\mathbf{-18.9}$ & $[-26.9, -10.8]$ \\
      & \lang{tur} & $83.6$ & $\mathbf{-8.2}$ & $[-15.5, -0.9]$ & $85.5$ & $-6.9$ & $[-14.0, +0.0]$ \\
    \midrule
    \multirow{4}{*}{Search}
      & \lang{eng} & $92.5$ & -- & & $97.5$ & -- & \\
      & \lang{ita} & $85.6$ & $-6.9$ & $[-13.9, +0.0]$ & $88.8$ & $\mathbf{-8.8}$ & $[-14.8, -3.3]$ \\
      & \lang{jpn} & $81.9$ & $\mathbf{-10.6}$ & $[-18.0, -3.3]$ & $87.5$ & $\mathbf{-10.0}$ & $[-16.2, -4.3]$ \\
      & \lang{tur} & $80.6$ & $\mathbf{-11.9}$ & $[-19.4, -4.4]$ & $87.5$ & $\mathbf{-10.0}$ & $[-16.2, -4.3]$ \\
    \midrule
    \multirow{4}{*}{Adaptability}
      & \lang{eng} & $76.7$ & -- & & $75.5$ & -- & \\
      & \lang{ita} & $71.1$ & $-5.7$ & $[-15.2, +4.0]$ & $74.8$ & $-0.6$ & $[-10.1, +8.8]$ \\
      & \lang{jpn} & $66.0$ & $\mathbf{-10.7}$ & $[-20.4, -0.8]$ & $73.6$ & $-1.9$ & $[-11.4, +7.7]$ \\
      & \lang{tur} & $76.7$ & $0.0$ & $[-9.3, +9.3]$ & $79.9$ & $+4.4$ & $[-4.8, +13.5]$ \\
    \midrule
    \multirow{4}{*}{Ambiguity}
      & \lang{eng} & $75.5$ & -- & & $78.0$ & -- & \\
      & \lang{ita} & $66.0$ & $-9.4$ & $[-19.2, +0.6]$ & $66.7$ & $\mathbf{-11.3}$ & $[-20.9, -1.5]$ \\
      & \lang{jpn} & $61.6$ & $\mathbf{-13.8}$ & $[-23.7, -3.6]$ & $63.9$ & $\mathbf{-14.1}$ & $[-23.7, -4.1]$ \\
      & \lang{tur} & $72.8$ & $-2.7$ & $[-12.3, +6.9]$ & $74.7$ & $-3.3$ & $[-12.6, +6.1]$ \\
    \bottomrule
  \end{tabular}
  \caption{\claude{} reasoning-effort ladder at the two highest tested settings (\texttt{high}, \texttt{xhigh}): per language, the \passk{3} and the target-minus-English marginal gap with its $95\%$ confidence interval, grouped by capability with the English reference (\lang{eng}) as the first row of each block (no gap, ``--'').
  CIs are Wilson--Newcombe intervals on the difference of two independent proportions, and gaps whose CI excludes zero are in \textbf{bold} ($7/12$ at either setting).}
  \label{tab:effort}
\end{table*}

\clearpage
\section{Trajectory triage protocol}
\label{app:triage-skill}

The automatic estimation of Section~\ref{sec:automaticgap} is produced by a
reusable diagnostic procedure that reads the evaluation harness's per-run
artefacts directly and classifies each failure under the taxonomy of
Section~\ref{sec:taxonomy}. We describe it here in full for reproducibility,
together with the infrastructure correction, the stratified reweighting, the
whole-dataset extrapolation, and the human validation that back the numbers in
the main text.

\paragraph{Inputs.}
For every (scenario, language, run) the harness persists: (i)~a \emph{verdict}
record with the pass/fail judgment and the failing check(s); (ii)~an
\emph{environment action log} of the agent's tool calls, from which the graded
\emph{write} actions, the user task, and the final answer are recovered;
(iii)~the raw \emph{agent trajectory} (LLM calls and reasoning); and, in newer
dumps, (iv)~a \emph{judge log} carrying, per oracle event, the LLM-as-Judge
rationale and the \emph{oracle reference} (the exact expected content). Runs are
indexed once; the diagnostic agents then address individual run directories
rather than scanning the corpus.

\paragraph{Phase 1 --- worklist construction.}
From the index we pair each target run with the English run(s) of the same
scenario. In \emph{regression mode} we retain scenarios where English passes and
the target does not; in \emph{failures mode} (no baseline) we retain failing runs
directly. Each work item records the run directories to compare and a
pre-computed failure family (check type, tool-count mismatch, loop/timeout,
termination), which supplies a coarse prior before any trajectory is read.

\paragraph{Phase 2 --- per-unit diagnosis.}
One agent processes one scenario (regression mode) or one failing run (the finer,
stratified mode used for the gap decomposition). It applies the taxonomy
\emph{top-down}: infrastructure first (a run terminated before a gradable attempt
is never blamed on the model), then translation defect, verifier artefact, and
agent failure, with inconclusive reserved for genuinely under-determined cases.
The English pass run is the oracle the dataset does not otherwise provide: if the
target agent took the same correct action yet the verifier rejected it, the verdict
is a verifier artefact; if the translated inputs (prompt/universe/oracle) changed
the correct action, it is a translation defect; if the inputs are faithful and the
model still erred, it is an agent failure. Following the standard priority rule, a
translation defect that also induces a downstream model error is attributed to the
translation. Judge rejections---otherwise ambiguous between a verifier
artefact (K1) and a genuine wrong answer (A5)---are decided by comparing the
agent's produced value against the persisted oracle reference and judge rationale,
never from which check fired alone.

\paragraph{Phase 3 --- compilation and estimation.}
Per-unit verdicts are aggregated into the fault-side split, the
category$\times$capability and per-language breakdowns, and a sub-code
distribution. Two corrections make the aggregate gap-representative. First,
\emph{infrastructure removal} (below). Second, \emph{stratified reweighting}:
writing $R_{\ell,s}$ for the true number of infra-clean failing runs of language
$\ell$ in determinism stratum $s\in\{\mathrm{det},\mathrm{stoch}\}$ (obtained
exactly from the run index) and $\hat p_{\ell,s}(c)$ for the sampled share of
cause $c$ in that cell, the per-language composition is
\[
  \phi_\ell(c)=\frac{\sum_{s} R_{\ell,s}\,\hat p_{\ell,s}(c)}{\sum_{s} R_{\ell,s}},
  \qquad
  \phi(c)=\frac{\sum_{\ell,s} R_{\ell,s}\,\hat p_{\ell,s}(c)}{\sum_{\ell,s} R_{\ell,s}},
\]
with $\phi(c)$ the pooled estimate reported in Table~\ref{tab:auto-split}.
Uncertainty is estimated by resampling scenarios (not runs) with replacement
within each $(\ell,s)$ cell---a cluster bootstrap, $B=2000$---which propagates both
sampling variance and the correlation among runs of the same scenario.

\paragraph{Infrastructure removal.}
A run is labelled \emph{infrastructure} if it carries the harness termination
sentinel, a null verdict, or appears in the list of batch-failed
(language, capability, run) cells arising from per-shard cold starts. Such runs
persist a non-null but non-gradable outcome, so a naive reading would grade them
as agent errors; they account for $8.5\%$ of all runs and are excluded from every
attribution rate. The batch failures are strongly cell-localised (e.g.\ a single
cold-started shard can void a majority of one language--capability--run cell), so
they distort per-language scores unevenly if not removed; correcting them was the
largest single revision to earlier estimates.

\paragraph{Sampling.}
For the gap decomposition we triage a language- and capability-balanced sample of
infra-clean regressions in each determinism stratum ($\approx\!4$ deterministic
and $8$ stochastic scenarios per language$\times$capability across the ten target
languages), diagnosed at the granularity of individual failing runs. For the
whole-dataset audit (below) we additionally sample $\approx\!11$--$12$ scenarios
per language from each of the both-fail and partial$\to$fail cells. All samples
use a fixed seed; per-language stratum sizes $R_{\ell,s}$ are taken from the full
index, not the sample.

\paragraph{Whole-dataset extrapolation.}
The triage above conditions on English passing and is by construction blind to
scenarios that fail in \emph{both} languages---precisely where a translation
defect could deflate a target score without leaving a visible regression. To
bound contamination over the entire benchmark we cross-tabulate all $6{,}056$
English$\times$target scenario--language pairs by outcome (Table~\ref{tab:contingency})
and use the observation that a translation defect can only corrupt a score by
rendering the target \emph{unsolvable} (the target never passes). Any pair whose
target succeeds on $\geq\!1$ run is therefore translation-clean by construction
($74.5\%$ of pairs). The remaining ``target-never-solves'' pairs partition into
three cells---English-solves (pass$\to$fail), English-partial
(partial$\to$fail), and English-fails (fail$\to$fail)---each of which we triage;
the whole-dataset unsolvable-due-to-translation rate is their size-weighted
average. The translation-defect rate falls monotonically as English competence
drops ($59\%$, $16\%$, $10\%$ respectively), yielding the $6.4\%$ bound of
Section~\ref{sec:automaticgap} and showing the both-fail blind spot to be cleaner
than the visible gap.

\begin{table}[h]
\centering\small
\begin{tabular}{lcccc}
\toprule
& \multicolumn{3}{c}{\textbf{Target}} & \\
\cmidrule(lr){2-4}
\textbf{English} & pass & partial & fail & \textbf{row} \\
\midrule
pass    & 2199 & 981 & 435 & 3615 (59.7\%) \\
partial & 400  & 611 & 382 & 1393 (23.0\%) \\
fail    & 49   & 271 & 728 & 1048 (17.3\%) \\
\bottomrule
\end{tabular}
\caption{Outcome contingency over all $6{,}056$ English$\times$target
scenario--language pairs (\claude{}, infra-clean, $\geq\!2$ valid runs per side).
Pairs whose target passes on $\geq\!1$ run ($74.5\%$) are translation-clean by
construction. The three ``target never passes'' cells---pass$\to$fail (clean
regression, $59\%$ translation), partial$\to$fail ($16\%$), and fail$\to$fail
(the both-fail blind spot, $10\%$)---are each triaged; their size-weighted
translation rate is the $6.4\%$ whole-dataset bound.}
\label{tab:contingency}
\end{table}

\paragraph{Auditing the both-fail cell.}
Because both-fail scenarios have no passing baseline in either language, we do not
judge the agent there. Instead we run an \emph{input-faithfulness} audit: a
translation defect is a property of the inputs, so we compare the target prompt,
universe, and oracle reference against their English source and flag only
discrepancies that would change the correct action. This detects translation
contamination independently of any pass/fail signal. The audit finds a
translation-defect rate of $9.9\%$ (95\% CI $5.6$--$16.9$) in the both-fail cell,
clustered on a small number of oracle- and keyword-mistranslation bugs (a single
scenario accounted for defects in five languages).

\paragraph{Per-language decomposition.}
Table~\ref{tab:perlang} reports the reweighted composition and the resulting
model-driven share of the conditional gap by language. The genuine (agent) share
is largest for Turkish and Japanese and smallest for Portuguese-\ and
Spanish-heavy comparisons; the translation share is largest for Hindi and
Chinese. Per-language intervals are wide (each rests on $\approx\!15$ sampled
scenarios), so they are indicative; the pooled estimate (Table~\ref{tab:auto-split})
is the reliable quantity.

\begin{table}[h]
\centering\small
\begin{tabular}{lccccc}
\toprule
\textbf{Lang} & \textbf{Transl.} & \textbf{Verifier} & \textbf{Agent} & \textbf{Agent 95\% CI} & \textbf{model gap (pp)} \\
\midrule
tur & 16\% &  6\% & 77\% & [62, 90] & 20.9 \\
jpn & 30\% & 13\% & 57\% & [40, 74] & 17.8 \\
por & 26\% &  6\% & 68\% & [58, 78] & 15.2 \\
hin & 48\% &  3\% & 49\% & [31, 67] & 14.1 \\
fra & 35\% &  6\% & 60\% & [45, 75] & 12.6 \\
cmn & 45\% &  9\% & 46\% & [29, 64] & 12.5 \\
deu & 43\% &  8\% & 49\% & [35, 62] & 11.0 \\
spa & 24\% & 12\% & 63\% & [49, 78] & 10.9 \\
ita & 37\% & 23\% & 40\% & [25, 58] &  9.7 \\
ind & 38\% & 15\% & 47\% & [32, 63] &  8.9 \\
\bottomrule
\end{tabular}
\caption{Per-language reweighted gap composition and model-driven gap
(\claude{}). ``Model gap'' is the conditional target failure rate on
English-solved scenarios times the agent share. Per-language CIs are wide
($\approx\!15$ scenarios each); the pooled estimate is authoritative.}
\label{tab:perlang}
\end{table}

\paragraph{Reliability and validation.}
Each verdict carries a self-reported confidence ($76\%$ high on the primary
regression pass). As an external check, expert linguists independently
re-analysed a subset of scenarios (Section~\ref{sec:linguisticanalysis}); their
attributions on deterministic regressions agree with the automatic labels and
with the translation-dominated composition of that stratum, and the deep-dive
cases in Table~\ref{tab:merged-cases} were drawn from this validation. The main
residual uncertainty is that the automatic pass tends to \emph{under}-attribute to
the verifier (borderline judge cases are conservatively labelled
translation defects), so the translation share in Table~\ref{tab:auto-split} is a
mild upper bound and the verifier-artefact share a lower bound; the model-side
share, and the whole-dataset contamination bound, are unaffected.

\section{Non-Latin Script Errors} \label{script_analysis}

We investigate why performance on languages with non-Latin script were the lowest, despite these being some of the highest-resource languages (notably cmn). This gap to non-Latin languages is consistent across all models. Our initial hypothesis was that translation issues were the cause, but \emph{we find no translation/transliteration issues from benchmark construction and find all scenarios solvable}.

We use \textsc{Claude-5-Sonnet} to identify and categorise the agentic failures of \claude in cmn, hin, jpn. Across 939 scenarios across the three languages, 88 failures were script-related (9\%). In such failures, we categorise 35 (4\%) to be fully the agent's fault and 53 (5\%) to be the agent's mishandling of script inconsistencies in the environment data. 
The most common type of error is when the agent searches/filters on only one form of an entity (e.g. written in Chinese characters) and fails because the tool was expecting it in a different form (e.g. Latin characters). In many cases, the agent explicitly noticed both forms in its own reasoning, but still chose to match only one. This class of errors is similar to those identified by \citet{bandarkar2026largereasoningmodelsstruggle}. While cross-script settings add a layer of difficulty, a highly capable reasoning model is expected to navigate such situations more effectively (e.g. retrying the search with multiple forms). We conservatively estimate that this agent's lack of care in multi-script scenarios accounts for about a $2$\,pp drop in \passk{3} accuracy, about half of the gap to Latin languages like spa and ind. We anticipate that in real-world multilingual agentic settings, there exists as many, if not more, script inconsistencies in tools and data.

\section{Human Validation of Triage Labels}
\label{app:triage-validation}

To verify the automatic attribution (\S\ref{sec:automaticgap}) at scale, we had
native or fluent speakers re-adjudicate a sample of triaged failures through a
purpose-built review interface, independently of the automatic protocol.

\paragraph{Protocol.}
Each item pairs one failing target-language run with its automatic verdict. The
reviewer sees only the decision-relevant evidence---the English source task, its
translation, the oracle reference, the LLM-as-Judge rationale (where applicable),
and the agent's final answer---rather than the full trajectory. For every item the reviewer either \emph{confirms} the
assigned fault side, \emph{rejects} it and names the side they believe is correct,
or marks it \emph{undecidable} when the evidence is insufficient. The three sides
mirror the taxonomy of \S\ref{sec:taxonomy}: translation defect (T), verifier
artefact (K), and agent failure (A); infrastructure and inconclusive items are not
sampled. Items are drawn from both regression strata (deterministic \emph{gap} and
stochastic \emph{partfail}) and balanced across languages and capabilities.

\paragraph{Coverage.}
Five linguists contributed \textbf{149} items in
seven languages (spa, ita, hin, jpn, ind, por, cmn) and all four capabilities.
Excluding the 9 undecidable items, \textbf{128 of 140} adjudications confirmed the
automatic label---\textbf{91.4\%} agreement (95\% Wilson CI $85.6$--$95.0$).
Agreement is comparable across fault sides (Table~\ref{tab:val-side}) and strata
(gap $92.9\%$, partfail $88.1\%$). Per-language agreement (Table~\ref{tab:val-lang}) is uniformly
high except for Japanese ($68\%$), where reviewers reattributed several
agent-failure labels to translation defects---a small, script-specific pocket we
flag for follow-up rather than a systematic bias.

\begin{table}[h]
\centering\small
\begin{tabular}{lccc}
\toprule
\textbf{Fault side} & \textbf{Adjudicated} & \textbf{Confirmed} & \textbf{Agreement} \\
\midrule
Translation defect (T) & 22  & 19  & 86.4\% \\
Verifier artefact (K)  & 27  & 25  & 92.6\% \\
Agent failure (A)      & 91  & 84  & 92.3\% \\
\midrule
\textbf{All}           & \textbf{140} & \textbf{128} & \textbf{91.4\%} \\
\bottomrule
\end{tabular}
\caption{Human re-adjudication of automatic triage labels by fault side
(undecidable items excluded). Agreement $=$ confirmed $/$ adjudicated.}
\label{tab:val-side}
\end{table}

\begin{table}[h]
\centering\small
\begin{tabular}{lccc}
\toprule
\textbf{Language} & \textbf{Adjudicated} & \textbf{Confirmed} & \textbf{Agreement} \\
\midrule
Spanish    & 29 & 27 & 93.1\% \\
Italian    & 26 & 25 & 96.2\% \\
Hindi      & 23 & 23 & 100\%  \\
Japanese   & 22 & 15 & 68.2\% \\
Indonesian & 22 & 22 & 100\%  \\
Portuguese & 10 & 9  & 90.0\% \\
Mandarin   & 8  & 7  & 87.5\% \\
\midrule
\textbf{All} & \textbf{140} & \textbf{128} & \textbf{91.4\%} \\
\bottomrule
\end{tabular}
\caption{Human re-adjudication agreement by language (undecidable items excluded).}
\label{tab:val-lang}
\end{table}

\end{document}